%% file: main.tex
\documentclass[lettersize,journal]{IEEEtran}
\usepackage[caption=false]{subfig}
\usepackage{adjustbox}
\usepackage{silence}
\usepackage{array}
\newcolumntype{x}[1]{>{\centering\arraybackslash\hspace{0pt}}p{#1}}
\usepackage{graphicx}
\usepackage[utf8]{inputenc} 
\usepackage[T1]{fontenc}    
\usepackage{hyperref}       
\usepackage{url}            
\usepackage{booktabs}       
\usepackage{amsfonts}       
\usepackage{nicefrac}       
\usepackage{microtype}      
\usepackage{xcolor}         
\usepackage{float}
\usepackage{algorithm}
\usepackage{algorithmic}
\usepackage{wrapfig}
\usepackage{supertabular,booktabs}

\usepackage[misc,geometry]{ifsym}

\usepackage{pythonhighlight}
\usepackage[center]{caption}
\usepackage{listings}
\usepackage{longtable}
\usepackage{amsmath,amscd,amsbsy,amssymb,latexsym,url,bm,amsthm}
\newtheorem*{remark}{Remark}

\usepackage{hyperref}
\usepackage{balance}

\ifCLASSOPTIONcompsoc
  \usepackage[nocompress]{cite}
\else
  \usepackage{cite}
\fi
\ifCLASSINFOpdf
\else
\fi
\begin{document}

\title{Master-slave Deep Architecture for Top-$K$ Multi-armed Bandits  with Non-linear  Bandit Feedback  and Diversity Constraints}
\author{
Hanchi Huang$^*$ \thanks{* Work done at Tencent.}, Li Shen, Deheng Ye, Wei Liu,~\IEEEmembership{Fellow,~IEEE} \\
\thanks{This work is supported by Science and Technology Innovation 2030 –“Brain Science and Brain-like Research” key Project (No. 2021ZD0201405).}
\thanks{Hanchi Huang is with Nanyang Technological University, Singapore (hhuang036@e.ntu.edu.sg). }
\thanks{Li. Shen is with JD Explore Academy, Beijing, China (mathshenli@gmail.com). }
\thanks{Deheng Ye and Wei Liu are with Tencent, Shenzhen, China (dericye@tencent.com, wl2223@columbia.edu).}
\thanks{Li Shen is the corresponding author.}
}

\markboth{Journal of \LaTeX\ Class Files,~Vol.~14, No.~8, August~2021}%
{Shell \MakeLowercase{\textit{et al.}}: A Sample Article Using IEEEtran.cls for IEEE Journals}


\maketitle 
\begin{abstract}

We propose a novel master-slave architecture to solve the top-$K$ combinatorial multi-armed bandits problem with non-linear bandit feedback and diversity constraints, which, to the best of our knowledge, is the first combinatorial bandits setting considering diversity constraints under bandit feedback. Specifically, to efficiently explore the  combinatorial and constrained action space, we introduce six slave models with distinguished merits to generate diversified samples well balancing rewards and constraints as well as efficiency. Moreover, we propose teacher learning based optimization and the policy co-training technique to boost the performance of the multiple slave models. The master model then collects the elite  samples provided by the slave models and selects the best sample estimated by a neural contextual UCB-based network to make a decision with a  trade-off between exploration and exploitation.  Thanks to the elaborate design of slave models, the co-training mechanism among slave models, and the novel interactions between the master and slave models, our approach significantly surpasses existing state-of-the-art algorithms in both synthetic and real datasets for recommendation tasks.   The code is available at: \url{https://github.com/huanghanchi/Master-slave-Algorithm-for-Top-K-Bandits}.
\end{abstract}


\input{1_1_introduction}
\input{1_2_relatedwork}

\input{2_method}

\input{3_experiment}

\input{4_conclusion}

\bibliography{sample-base}
\bibliographystyle{abbrv}


\vspace{-10 mm}

\clearpage
\include{5_appendix.tex}
\end{document}

%% file: 1_1_introduction.tex
\section{Introduction}
\IEEEPARstart{M}{ulti-armed} bandit (MAB) is a vital framework to model the trade-off between exploitation and exploration when playing multi-armed bandit machines. It has been extensively studied in many fields \cite{ditzler2017sequential,santosh2022multiarmed}, such as clinical trials, financial portfolio design, recommendation systems, etc. 

Top-$K$ combinatorial MAB (CMAB) \cite{cao2015top} is a special variant in MAB where each action is a subset of arms with size $K$ out of $L$ arms.
Generally, the expected aggregated reward of the action is the sum of rewards of all the selected arms. As to the observability of the feedback, most works use semi-bandit feedback, which assumes the player can observe the rewards for each selected arm \cite{zimmert2019beating}.
In the meantime,  bandit feedback \cite{kuroki2020polynomial}, that the player can only observe the aggregated reward of the whole action instead of  rewards for each arm  in the action, has also been used. Since in recommendation systems, it is costly to obtain a user's rewards for each selected item,   the semi-bandit feedback might  be infeasible.
In this paper, we focus on top-$K$ CMAB under bandit feedback.

Besides bandit feedback, we also consider the non-linear rewards and diversity constraints. 
First, it might be not practical in all real recommendation systems to assume a  user's aggregated reward for an action is the linear function of rewards for each selected arm \cite{rejwan2020top}, which motivates us to consider the top-$K$ CMAB with non-linear rewards.
Second, for real top-$K$ recommendations where an item is regarded as an arm and the recommended action is a set with $K$ items, a user may propose many demands on items in an action, which can impact the ultimate feedback to those items. 
When considering a user's demands, it is more practical to introduce extra constraints to portray the interaction among items \cite{ parameswaran2011recommendation}, such as diversity constraints \cite{gong2019exact} where diversification is one of the leading topics in recommendation systems and helps Taobao \cite{gong2019exact} obtain huge profits. Thus, we also try to explore the top-$K$ CMAB setting with diversity constraints.

Solving the above online constrained combinatorial optimization problems (O-CCOPs) can face many challenges such as huge action space, great difficulty to satisfy constraints, and handling the reward-constraint   trade-off. Since we do not know much environmental information, we cannot use traditional optimization solvers to solve the above problems and need to resort to deep neural networks. However, normal 
deep learning based methods are poor at constraint-handling and are time-consuming for training, so we modify and co-train solvers and deep (reinforcement) learning methods to tackle the above challenges. Below, we summarize our approach and contributions.

\emph{Our approach.}\ 
We design a novel master-slave hierarchical architecture
that leverages the cooperation of learning, sampling, and traditional optimization methods,
to efficiently explore the combinatorial and constrained action space.
For the master model, we utilize a neural contextual UCB-based network  to estimate the user feedback, provide surrogate rewards to train slave samplers, and estimate  samples collected from slave models to make a decision. 
Apart from the master model, we deliberately design multiple slave models, each of which has its  distinguishing merit to deal with the  proposed setting:
\textbf{(i)} 
Solver sampler finds the solutions that strictly satisfy all constraints but may not own high rewards.
\textbf{(ii)}  Since owning low constraint
violation rates does not mean owning high rewards, we adopt the primal-dual Wolpertinger sampler which
can efficiently search in the combinatorial space by designing proto-action and balance the reward-constraint
tradeoff by primal-dual-based reward shaping. 
\textbf{(iii)} G2anet sampler makes up for the pity that the above samplers do not thoroughly study the influence of constraints on each pair of variables.
\textbf{(iv)} Both Wolpertinger and G2anet require time-consuming back-propagation. In contrast, the CEM-PPO sampler requires much less of it, can be naturally used for various O-CCOPs, is easier to scale in distributed environments, and has fewer hyper-parameters.
\textbf{(v)} Teacher-student sampler promotes samplers to learn from each other and generate more diversified and elite samples. 
During the sample generating process, we also propose  policy co-training to avoid local optimal solutions and achieve mutual benefit among  slave models.

\emph{Our contributions.}\   
(i) We introduce a master-slave hierarchical algorithm to cope with the first top-$K$ CMAB setting under non-linear bandit feedback and diversity constraints. 
(ii) We propose multiple slave models, each of which has its  distinguishing merit to produce diversified candidate solutions for solving constrained optimization problems.  An improved gumbel top-$K$ sampling trick is embedded in slave models to guarantee their end-to-end training.
(iii) Due to our elaborate design of interactions between master and slave models, as well as the learning from demonstrations and the sample sharing  mechanisms (two approaches of policy co-training) among slave models, our architecture achieves significant results in experiments.

%% file: 1_2_relatedwork.tex
\section{Related Work}
\emph{Machine learning-based recommendation system.} A recommendation system is a  tool that analyzes users' interests and preferences to recommend products, services, or information that they may find interesting. Nowadays, it is widely used in e-commerce websites, video platforms, news portals, and other applications with the main goal of providing personalized recommendations to enhance users' experience.
In recent years, machine learning-based
recommendation systems \cite{luo2015nonnegative,zhang2018cross,wu20211,ding2022causal} have made progress in many directions, such as: (i) 
application of deep learning: Deep learning models have the capability to automatically learn high-dimensional feature representations, enabling them to capture complex relationships between user interests and item features. This significantly improves the effectiveness of recommendations. Typical models include Wide\&Deep \cite{cheng2016wide}, DeepFM \cite{DBLP:conf/ijcai/GuoTYLH17}, and Deep Crossing \cite{shan2016deep};
(ii) multi-task learning: Recommendation systems often involve multiple related tasks, such as click-through rate (CTR) estimation, conversion rate (CVR) estimation, and dwell time prediction. Multi-task learning allows parameter sharing and mutual enhancement among different tasks, thereby improving recommendation performance. Typical models of multi-task learning in recommendation systems include MMoE \cite{ma2018modeling} and PLE \cite{tang2020progressive};
(iii) cross-domain recommendation: Cross-domain recommendation aims to leverage user interests in one domain to recommend items in another domain, achieving knowledge transfer between domains. This requires models to learn commonalities and differences across domains to enable personalized recommendations. Typical works include SAR-Net \cite{shen2021sar};
(iv) diversity recommendation: Diversity recommendation aims to avoid recommending only popular items and instead recommends a variety of items to satisfy different users' interests and preferences, thereby enhancing users' experience. DPPNET \cite{mariet2019dppnet} is a common example of a diversity recommendation model that uses deep neural networks  to approximate diverse determinantal point processes (DPPs);
(v) reinforcement learning: Reinforcement learning enables automatic learning of optimal strategies in complex environments. Its application in recommendation systems is still immature but holds great potential. Typical works include deep reinforcement learning for news recommendation \cite{zheng2018drn}.
In summary, machine learning-based recommendation systems are rapidly evolving to address various technical challenges in practical recommendation processes. These techniques aim to continually improve the accuracy and diversity of recommendations and optimize users' experience.

\emph{Combinatorial multi-armed bandits (CMAB).}\ 
CMAB has been widely studied with various applications, such as graph routing, online recommendation, influence maximization, etc.
Most works on CMAB assume the player could observe the rewards for each arm in an action \cite{VermaHRS19}.  In this setting, Bubeck et al. \cite{bubeck2012towards} introduced Online Stochastic Mirror Descent (OSMD)  
with minimax optimal regret bounds in two
action sets, the hypercube and the Euclidean ball. Later on, Kuroki et al. \cite{kuroki2020polynomial} proposed a polynomial-time bandit algorithm for a $0-1$ quadratic programming problem with high  statistical efficiency. Recently, Rejwan and Mansour \cite{rejwan2020top} proposed combinatorial successive accepts and rejects based on a sampling method  utilizing Hadamard matrices to estimate arms with a small number of samples. 
All these above works considered the aggregated  reward  as the sum of individual rewards. 
On the contrary, in \cite{agarwal2018regret}, the aggregated reward is generalized to the weighted sum and the non-linear function of  each individual reward in an action, respectively. 
Lin et al. \cite{lin2014combinatorial}  presented the global confidence bound algorithm, 
which borrows ideas from both the estimation of the small base arm in CMAB and the finite partial monitoring games to deal with  the issue of limited  feedback. 
Different from \cite{lin2014combinatorial},
Agarwal et al. \cite{agarwal2018regret} designed a  divide-and-conquer based strategy called CMAB-SM, which has been proven to be both space-efficient and computationally efficient.
However, all the above works do not consider imposing any constraint on selected arms 
in CMAB.

\emph{Neural combinatorial optimization.}\
Recently, pointer network \cite{bello2016neural} and graph neural networks have been employed to end-to-end decision and representation learning for combinatorial optimization problems (COPs) \cite{gasse2019exact,wang2023towards}.  
There are also several works focusing on the enhancement of traditional  methods  for COPs \cite{BalcanDSV18,EtheveABJK20}.
Among these, machine learning and reinforcement learning (RL) techniques are utilized under the branch-and-bound and cutting planes framework for branch decision, node selection, etc. 
However, the above methods do not tightly apply to online constrained COPs where constraint handling in online   learning adds much more difficulties, and 
CMAB with hard constraints is, therefore, still an open challenge, which is just the focus of this paper.

%% file: 2_method.tex
\section{Methodology}

In this section, we describe our master-slave model in detail. 
We start by introducing the top-$K$ CMAB under non-linear bandit feedback and diversity constraints. 

\subsection{Problem setup}
 	 

Top-$K$ CMAB  under bandit feedback assumes there are $L$ arms in the arm set $\mathcal{D}=[L]$. At  time $t=1, 2, \ldots$, the player chooses an action set $a_t$ with $K$ arms $\{a_{t,1},\cdots,a_{t,K}\}$ from  $\mathcal{D}$  and only receives an aggregated feedback $r_t =f(a_t)+\epsilon_t$ from the environment, where $\epsilon_t$ is a $1$-sub-Gaussian noise conditioned on $a_1,a_2,\cdots,a_{t-1}$ satisfying $\mathbb{E}\epsilon_t=0$. The goal of
top-$K$ CMAB is to solve 
\begin{align}\label{top-k-cmab}
 \underset{a_t}{\max} \  \sum_{t=1}^{T} f(a_t), s.t.\ a_t\in \mathcal{D},\ Card(a_t)=K,
\end{align}
where $T$ is the number of recommendation rounds and  $Card(a_t)$ represents the number of non-zero elements in $a_t$.
Here we allow  $f$ to be  non-linear   on $a_t$ and require the following diversity constraints. 

Assume each arm is paired with a feature vector. Then the diversity constraints refer to that normalized edit distance (NED) \cite{marzal1993computation,gong2019exact} of any two arms’ feature vectors should be larger than a threshold $\tau$ as much as possible. To simplify the formulation, we denote the  action at each round $t$ as a binary vector $A_t=\{A_{t,1},\cdots,A_{t,L}\}\in \mathcal{A}_0=\{b \in\{0,1\}^{L} \mid \sum_{i=1}^{L} b_{i}=K\}$.
And we assume that if there exist arm $i$ and arm $j$ whose feature vectors' NED \cite{gong2019exact} is smaller than $\tau$, then it should hold that $A_{t,i}+A_{t,j}\leq 1$, which means these two similar arms should not co-exist in $A_t$. In conclusion, the specific formula for diversity constraint can be represented as follows:
\begin{align*}
 \underset{a_t}{\max} \quad & \sum_{t=1}^{T} f(a_t) \\
s.t. \quad & a_t\in \mathcal{D}, Card(a_t)=K; \\
& A_{t,i}+A_{t,j}\leq 1, \text{ if NED}(i,j) < \tau. \\
\end{align*}

In addition, we define the reward and constraint violation rate of $A_t$   as  $r_t=f (a_t)+\epsilon_t=h(A_t)+\epsilon_t$ and  $c_t = \frac{n_t}{M}$, where $h$ is a non-linear feedback function on $A_t$ instead of $a_t$, $n_t$ is the number of violated diversity constraints 
for action $A_t$, and $M$ is the number of all the  constraints.
The composite evaluation metric is naturally $r_t-\lambda c_t$, where $\lambda$ is a trade-off parameter between the reward and constraint violation term and is determined via specific circumstances\footnote{In experiments, our method is proved to enjoy both the best $r_t$ and the best $c_t$. Therefore for any composite metric that is monotonous to $r_t$  and $c_t$, our method also performs   best 
regarding that composite metric.}.



In brief, the goal of our setting,
top-$K$ CMAB with non-linear bandit feedback and diversity constraints, is to solve the following optimization problem:
\begin{gather}
 \max_{A_{1},A_{2},\cdots,A_{T}\in \mathcal{A}_0}
 \left\{\sum_{t=1}^{T} h(A_t),\ -\sum_{t=1}^{T} c_t\right\}. \label{eq:top-k-cmab-diversity}
\end{gather}
 This is a multi-objective optimization problem with a huge combinatorial action space.
Compared with classic top-$K$ CMAB \eqref{top-k-cmab}, the non-linear rewards and the additional requirement to minimize $\sum_{t=1}^{T} c_t$ bring a new challenge since it seems not easy to balance the reward-constraint trade-off in online learning scenarios where too many constraints exist. In this work, we integrate a deep reinforcement learning approach with a traditional optimization method  to tackle this challenging problem \eqref{eq:top-k-cmab-diversity}.

\subsection{Master-slave algorithm}



Since there are few algorithms to deal with non-linear rewards
in top-$K$ CMAB, we novelly convert our problem to contextual MAB with existing neural network based methods with a nearly-optimal guarantee and utilize a customized master-slave architecture to handle the differences between CMAB and contextual MAB. Specifically, we represent the action at round $t$ as a 
vector $A_t$ in contextual MAB rather than a set $a_t$
in CMAB. A neural contextual UCB-based network (NeuralUCB) \cite{zhou2019neural} is, then, utilized to 
estimate the item vector $A_t$. Here note that the 
action space of contextual bandits at each round is of fixed
and very limited size while   our problem owns a huge combinatorial action space which makes the evaluation for all candidate actions very expensive. To cope with this issue, we construct two types of models: one type  for action evaluation  and the other type   to 
select the most elite candidate actions of fixed size as the action space in contextual MAB. 

\begin{algorithm}[tb]
\flushleft
\small
\caption{Master-slave architecture} 
\label{alg:Master-slave}
\begin{algorithmic}[1]
    \STATE 
\textbf{Subproblem 1 (hard constraints):}
    $ {\max}_{A_{t}, t\in [T]} \left\{\sum_{t=1}^{T} h(A_t) \ {\rm s.t.\ }   A_{t}  {\rm \ satisfies\ all\ constraints}\right\}$.
    
   \STATE   \textbf{Subproblem 2 (soft constraints):}  $\max_{A_{t},t\in [T]} \left\{\sum_{t=1}^{T} h(A_t) -\lambda\sum_{t=1}^{T} c_t {\rm\ s.t.\ } A_{t}\in \mathcal{A}_0\right\}$.
    \FOR{\texttt{$t=1,2,\cdots$}}
   \STATE The Gurobi solver sampler solves \textbf{Subproblem 1} and the other $5$ samplers solve \textbf{Subproblem 2} 
   \STATE  
   When generating elite samples,
    policy co-training techniques are utilized for mutual benefits of slaves
    \STATE
    The master model evaluates all elite samples from the slave samplers and recommends the best one
   \STATE The master receives the user feedback and updates the estimation of the user's preference via NeuralUCB
    \ENDFOR
   
\end{algorithmic} 
\end{algorithm}
\begin{figure}
    \centering
    \includegraphics[width=\linewidth]{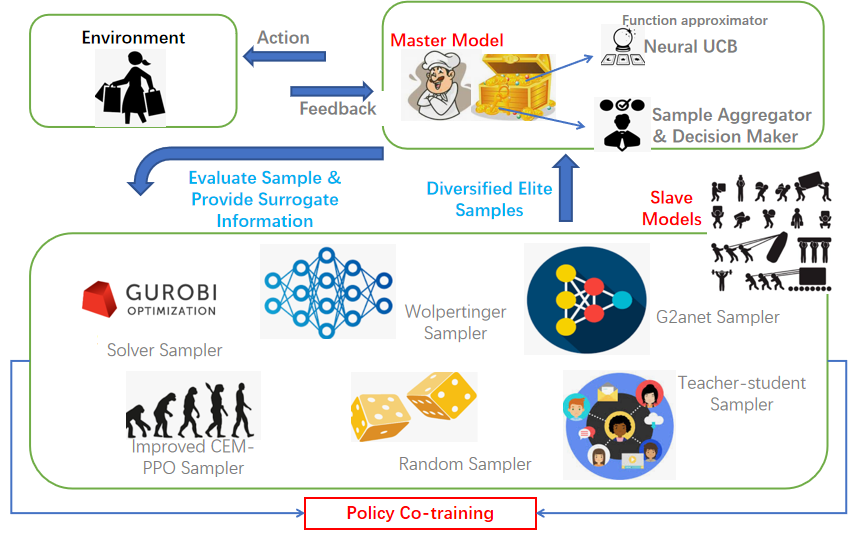}
    \caption{The pipeline of the master-slave algorithm.}
    \label{fig:master-slave structure}
    \vspace{-0.6cm}
\end{figure} 

As shown in Figure \ref{fig:master-slave structure} and Algorithm \ref{alg:Master-slave},  a master-slave architecture  containing a master model and six slave models is designed to interact with the environment.
At each round, the master model performs supervised learning with NeuralUCB \cite{zhou2019neural} based on the historical action-feedback interaction data  with the environment  to estimate the  environmental feedback function $h$. Then, the master model collects and evaluates  elite samples from six slave models to decide the action and receives the feedback  at round $t$.
The slave models, however, rely on surrogate feedback estimated by NeuralUCB to train themselves and provide diversified elite samples  for the master  to make a decision.
During the training, a policy co-training  technique is proposed to boost the performance of slave models. 

In our implementation, for the master,   we update   NeuralUCB  at each round. However for  slave samplers,
at the first $2L$ rounds, only the random sampler is utilized for enough exploration. When $t>2L$, all six samplers will be employed to provide  elite samples at each round and we update the samplers  every $f_{in}$ rounds. During the master-slave interactions, we can adjust the number of elite samples that each sampler should provide at each round according to the quality and diversity of  samples that each sampler provides in history. Besides, the interaction frequency is also adjustable to reduce the computational cost. For example, some samplers require time-consuming back-propagation and are sample-inefficient. For these samplers, we could decrease their interaction frequency with the master properly.


To sum up,  our master-slave hierarchical architecture with  various slave models owning different advantages and the co-training mechanism among slave models ensures the efficient exploration in the huge combinatorial and constrained space and thus the high quality of the action  selected from elite samples. 
Moreover,  through the reward-constraint trade-off by the refined Gurobi solver sampler, the powerful searchability in the combinatorial action space by the Wolpertinger sampler with customized prioritized experience replay,  the computational efficiency of the evolutionary algorithm in the improved CEM-PPO sampler, and the relationship inference capability of the  graph neural network G2Anet in a multi-agent environment, our architecture achieves the ingenious fusion of traditional optimization methods and deep (reinforcement) learning  methods in solving  online constrained COPs.
Next, we describe our master and slave models and their interactive mechanism in detail.

\subsection{Master model}
The master model acts as a feedback estimator and the ultimate decision maker.
It employs a neural contextual UCB-based network (NeuralUCB) to approximate  the environmental feedback function by performing supervised learning on existing data and designs a composite score based on the estimation by NeuralUCB  to evaluate  elite samples collected from various slaves and make the ultimate decision from the elite samples. The NeuralUCB in the master model also provides the first-order  and second-order extracted information for the solver slave sampler   and provides surrogate feedback to train networks in slave models for unlimited episodes. 

\textbf{Neural contextual UCB-based network.}\ 
Neural contextual UCB-based network (NeuralUCB) is  proposed in \cite{zhou2019neural} to deal with non-linear rewards in contextual bandits and enjoys the nearly optimal regret order  in
scenarios with non-linear rewards. Since we regard the action as a vector instead of a set, NeuralUCB can also adapt to our top-$K$ setting.  It can be regarded as a neural non-linearization of   Linear UCB   in \cite{chu2011contextual} and approximates $h$ by a fully connected neural network $h'(x;\theta)$ where $\theta$ is the network weights of NeuralUCB trained on data $(A_1,r_1),(A_2,r_2),\cdots,(A_{t-1},r_{t-1})$. To guarantee the decision with a certain amount of exploration, at each round $t$, NeuralUCB adds a UCB-based  neuralized confidence radius to $h'(x;\theta)$ to  
form $U_{t,x}$, the optimistic estimation of $h$.

\textbf{Elite sample aggregator and evaluator.}\ 
With the estimation by NeuralUCB, the master model constructs an overall evaluation metric for each elite sample  $A_t$ collected from slave models:
$Score (A_t)= U_{t,\hat{A}_t}-\lambda c(A_t),$
where $\hat{A}_t = A_t/\|A_t\|$ is the normalized version of $A_t$ since NeuralUCB takes normalized vectors as inputs,  and $c(A_t)$ is the constraint violation rate.
After collecting   elite samples from  slave models as an action candidate pool, the master model selects the sample with the highest $Score$ as the recommendation action  and receives a user's feedback.                     
\subsection{Slave models}

There are six diversified slave models to provide elite samples with various degrees of the trade-off between $r_t$ and $c_t$  and to generate  samples with the highest quality as much as possible. Next, we describe the  co-training mechanisms of six slave models and each slave model  in detail.

\subsubsection{Co-training of slave models}
\textbf{Learning from demonstrations.}
Individual training of slave models may stick to local optimal solutions and it is rather difficult for the overall performance to significantly outweigh the best individual performance among all  slave models.
Therefore to avoid local optimal solutions and achieve mutual benefit among slave models, below we utilize a policy co-training technique based on learning from demonstrations, which applies supervised learning 
by taking the demonstration trajectories as ground truth \cite{hussein2017imitation,gong2019exact}: 

(i)  First store  $A^{\star}$, the global elite sample with the highest estimated composite score in the recent $L_2$ rounds, and regard it as the one-step demonstration trajectory.

(ii) Each time when a slave model $\mathcal{S}$ converges to a local optimal solution, we define a  cross-entropy-like loss between $A^{\star}$ and $RV$, the real-valued output of $\mathcal{S}$,   as $\ell(RV|\Theta,A^{\star})=
\frac{1}{L} \sum_{i\in [L]}\!-\![A^{\star}_{i}\log (RV_{i})+(1\!-\!A^{\star}_{i})\log (1\!-\!RV_{i})]$
where $\Theta$ is  network parameters of  $\mathcal{S}$.

(iii) Run the neural network of  $\mathcal{S}$ several times to minimize  $\ell(RV|\theta,A^{\star})$  and then restore the normal training process for $\mathcal{S}$. 

Note that for simplicity, we do not adopt top-$K$ off-policy correction  \cite{chen2019top} and adversarial training \cite{song2020co} in this part and leave deeper policy co-training  for multiple slave models as the future work.

\textbf{Sample sharing mechanism.}
(i) Always keep the $length_{epoch}$ historical elite samples with the highest surrogate score. The improved CEM-PPO sampler, which will be described later on, partially uses these elite samples to update parameters;
(ii) Always keep the latest $length_{epoch}$ actions actually recommended by the master. The primal-dual Wolpertinger and the G2ANet samplers, which will be described later on, use these  to update their critic networks.

\subsubsection{Solver sampler}
Since it might be rather difficult to stably trade-off the complex constraint term and the objective term in the loss with regard to deep (reinforcement) learning samplers, we directly adopt a mathematical programming solver to keep a balance between $r_t$ and $c_t$. We realize this goal by
extracting the first-order and second-order information from  NeuralUCB in our master model and taking this information and all the constraints as the inputs for Gurobi solver in \cite{optimization2014inc} to obtain elite samples which are proved to be leading components, i.e. providing nearly the most best elite samples, among samples from all slave samplers in the experimental evaluation.

\textbf{First-order information extractor.}\ 
To extract the first-order information of NeuralUCB,
we linearize $h^{\prime}$, the estimation of the experimental feedback function $h$, by taking the $L$ columns of the unit matrix $I^{L\times L}$ as the inputs of NeuralUCB, obtaining $L$ outputs $\{b_i\}_{i=1}^{L}$, and then formulating an integer programming (IP) problem:

\begin{equation}
    \begin{matrix}
 \underset{x}{\max} & \sum_{i=1}^{L}b_i x_i \\ 
 s.t. & \sum_{i=1}^{L}x_i=K; x_i\in \{0,1\}, \forall i\in [L];\\& x \text{ satisfies all diversity constraints.} 
\end{matrix} \label{eq:linear}
\end{equation}

Here, Problem  \eqref{eq:linear} is the approximate linearization  of the Subproblem 1. We apply Gurobi to solve   problem \eqref{eq:linear} to   obtain  elite sample $elite_1$  strictly satisfying all the constraints and enjoying high $h^{\prime}$ with high probability.

\textbf{Second-order information extractor.}
To extract the second-order information of NeuralUCB, we assume $h'(x)\approx x^T Q x +d^T x+e$ with $Q\in \mathbb{R}^{L\times L}$ and $Q=Q^T$,
and estimate $Q$ as follows: Assume $e_i$ is the $i$-th column of  the unit matrix $I_{L\times L}$.
Take $\{(e_i+e_j)/\sqrt{2}\}_{i,j\in [L],i\neq j}$ as the inputs of NeuralUCB and obtain outputs $\{o_{ij}\}_{i,j\in [L],i\neq j}$. Meanwhile, take 
 the zero vector as the inputs of NeuralUCB and obtain outputs  $e$. Then
$\forall i,j\in [L]$ with $i\neq j$, set $Q_{ij}=o_{ij}+\frac{\sqrt{2}}{4}(o_{ii}+o_{jj})-\frac{1+\sqrt{2}}{2}(b_i+b_j)+\frac{\sqrt{2}}{2}e$;  and $\forall i\in [L]$, set $Q_{ii}=\frac{o_{i i}-\sqrt{2}(b_i-e)-e}{2-\sqrt{2}}$.  And we formulate the following IP problem:

\begin{equation}
    \begin{matrix}
 \underset{x}{\max}  & x^T Q x  \\ 
 s.t. & \sum_{i=1}^{L}x_i=K;\\
& x_i\in \{0,1\}, \forall i\in [L];\\& x \text{ satisfies all diversity constraints.} \label{eq:5}
\end{matrix}
\end{equation}

Here, Problem  \eqref{eq:5} is the  quadratic approximation form  of the Subproblem 1. And we apply Gurobi to solve (\ref{eq:5}) to   obtain  elite sample $elite_2$.
\begin{remark}
The explanation of estimating $Q_{ij}$:
Since $h'(x)\approx x^T Q x +d^T x+e$ and $o_{ij}$ is the output of NeuralUCB with $(e_i+e_j)/\sqrt{2}$ as the input, it holds that  $o_{ij}\approx  (e_{i}+e_{j})^{T} Q(e_{i}+e_{j}) / 2+(e_i+e_j)/\sqrt{2})^{T}d+e$. 
Since $b_i=e_i^TQe_i+d^T e_i+e=Q_{ii}+d_i+e$, we have $o_{ij}\approx \frac{Q_{i i} +2Q_{ij }+Q_{jj }}{2}+\frac{(b_{i}-Q_{i i})+(b_{j}-Q_{j j})-2e}{\sqrt{2}}+e$. Therefore, $o_{i i} \approx 2Q_{i i}+\frac{2 b_{i}-2 Q_{i i}-2 e}{\sqrt{2}}+e$, and thus we have $Q_{ii}\approx\frac{o_{i i}-\sqrt{2}(b_i-e)-e}{2-\sqrt{2}}$. Furthermore, $o_{ij}\approx \frac{Q_{i i} +2Q_{ij }+Q_{jj }}{2}+\frac{(b_{i}-Q_{i i})+(b_{j}-Q_{j j})-2e}{\sqrt{2}}+e=
 \frac{Q_{i i} +2Q_{ij }+Q_{jj }}{2}+\frac{(b_{i}-\frac{o_{i i}-\sqrt{2}(b_i-e)-e}{2-\sqrt{2}})+(b_{j}-\frac{o_{jj}-\sqrt{2}(b_j-e)-e}{2-\sqrt{2}})-2e}{\sqrt{2}}+e=Q_{ij}-\frac{\sqrt{2}}{4}(o_{ii}+o_{jj})+\frac{1+\sqrt{2}}{2}(b_i+b_j)-\frac{\sqrt{2}}{2}e$.
 Therefore, it holds that $Q_{ij}\approx o_{ij}+\frac{\sqrt{2}}{4}(o_{ii}+o_{jj})-\frac{1+\sqrt{2}}{2}(b_i+b_j)+\frac{\sqrt{2}}{2}e$.

\end{remark}

\textbf{Beta-sampling perturber.}\ 
After obtaining two elite samples from the solver sampler, we utilize a beta-sampling perturber onto the two samples and obtain more diversified elite samples:
For $i\in \{1,2\}$,
 (i) set $elite_{new}=\{clip(elite_{i,j},\epsilon_0,1-\epsilon_0)\}_{j\in [L]}$;
    (ii) for $j\in [L]$, replace $elite_{new,j}$ with a random number sampled from $beta(elite_{new,j},1-elite_{new,j})$;
(iii) set the $K$ largest components in $elite_{new}$ as $1$ and the other components to be $0$.
Run the above procedure  several times and we can obtain several new elite samples.

\subsubsection{Primal-dual Wolpertinger sampler}

Although the solver sampler is excellent at constraint-handling, its performance is not guaranteed when the feedback function $h$ and the required constraints are not regular, and hence we need some general samplers to deal with more complex feedback and constraints. Considering this demand,
our primal-dual Wolpertinger Policy based sampler  integrates the primal-dual based reward constrained policy optimization algorithm  \cite{tessler2018reward}, scale-invariant of the reward and constraint terms,  with Wolpertinger Policy \cite{dulac2015deep}, a deep reinforcement learning  structure  which  constrains the network outputs onto the discrete action space by the $K_1$-nearest-neighbor technique. Besides this integration, we also make a series of our own  improvements.

\textbf{Wolpertinger architecture.}
The Wolpertinger architecture is based on the actor-critic structure and employs the Deep
Deterministic Policy Gradient algorithm \cite{lillicrap2015continuous} to train its parameters. At each round, the agent will receive a proto-action from the actor, which may not belong to the action space. Therefore, the agent then  finds $K_1$-nearest actions to the proto-action from the action space and selects the one with the highest Q value from the $K_1$ action candidates. 
In our implementation, after obtaining the proto-action, denoted as $PA$, we construct $PA^1$ by adding an  upper confidence term \cite{lattimore2020bandit} (according to the historical recommended times of each item, i.e., the number of rounds that an item was recommended historically) to each component of $PA$,  
rank all components in $PA^1$, and reform $PA^1$ by setting $PA^1_i=1$ if the original $PA^1_i$ belongs to the largest $K$ components in the original $PA^1$; otherwise, set $PA^1_i=0$. Then, we randomly switch two components in the new $PA^1$  several times to construct several action candidates. And next select several candidates with the highest values evaluated by the critic network as the elite samples.

\textbf{Reward constrained policy optimization.}
The   Reward Constrained Policy Optimization (RCPO) algorithm is introduced in \cite{tessler2018reward} for solving the constrained Markov decision processes  (CMDP), which can  handle more general constraints, does not need prior knowledge, and is scale-invariant with regard to the  reward and constraint terms.
Based on the MDP framework,
CMDP  introduces a penalty $ c(s, a)$, a constraint $C(s_{t})=F(c(s_{t}, a_{t}), \ldots, c(s_{T}, a_{T}))$, and a threshold $\alpha\in [0,1]$, where $s_t$ and $a_t$ are the state and action at round $t$, respectively.
Denote $J_{C}^{\pi}=\mathbb{E}_{s \sim \mu}^{\pi}[C(s)]$ as the expectation over the constraint,
where $\mu$ is the distribution  the initial state obeys.
The problem thus becomes:$
\max _{\pi \in \Pi} J_{R}^{\pi}=\mathbb{E}_{s \sim \mu}^{\pi}[\sum_{t=0}^{\infty} \gamma^{t} r_{t}]=\sum_{s \in S} \mu(s) V_{R}^{\pi}(s), \text { s.t. }  J_{C}^{\pi} \leq \alpha,$
where $V_{R}^{\pi}(s)=\mathbb{E}^{\pi}[\sum_{t} \gamma^{t} r(s_{t}, a_{t}) \mid s_{0}=s]$ and $S$ is the state space.

Given the above CMDP, \cite{tessler2018reward} deduced an  unconstrained problem: $\min _{\lambda_2 \geq 0} \max _{\theta} L(\lambda_2, \theta)=\min _{\lambda_2 \geq 0} \max _{\theta}[J_{R}^{\pi_{\theta}}-\lambda_2 \cdot(J_{C}^{\pi_{\theta}}-\alpha)].$

RCPO proposes a multi-time scale approach to this unconstrained problem: 
in the inner cycle, it applies a TD-critic to estimate an alternative discounted objective; in the  intermediate cycle, RCPO performs policy gradient to update the policy; in the outer cycle, it ascends on the previous constraint so as to update the reward-constraint trade-off coefficient, i.e., $\lambda_2$.  
  
  In our implementation, we borrow the $\lambda_2$ updating idea from RCPO and let $Surrogate(\cdot) -\lambda_2 c(\cdot)$ be the composite feedback for the Wolpertinger Policy, where $Surrogate(\cdot)$ is the surrogate evaluator provided by NeuralUCB in the master model, and $ c(\cdot)$ refers to   constraint violation rate of a vector in $\mathcal{A}_0$. 
  
As to the design of state space,  we initialize $s_0$ by the $L$-sized vector $\{\frac{1}{L}\}_{i=1}^{L}$ and set $s_{t+1}$ to be the normalization of the vector composed of the number of times each arm has been selected in the  elite samples provided by the primal-dual Wolpertinger sampler in the historical $[0, \left \lfloor \frac{t}{length_{epoch}} \right \rfloor length_{epoch}]$ rounds. Therefore, the transitions between states are smooth and the training process might be smoother. Also,
  we encourage the primal-dual Wolpertinger sampler to learn from other global elite samples of the highest quality  in the recent epoch by pushing these samples into the replay buffer, and  the designs of the corresponding state-action transitions   are very clear and correct thanks to the deliberate design of state space.
  
  Except for the above improvements, we elaborately customize the prioritized experience replay mechanism
 with action clustering and reward-constraint hierarchies, which is proven to boost the performance remarkably in   experiments. The experience replay mechanism goes as follows:
 
  (i) To promote an exact approximation of $h^{\prime}$ for the critic network and the diversity of samples generated by the policy network, we first cluster all samples in replay buffer into $20$ groups and then sample a certain proportion of samples from each group uniformly at random. By this sampling method, we collect $\frac{1}{3}$ batch of samples to update the primal-dual Wolpertinger sampler.
 
 (ii) To encourage the sampler to generate samples with high surrogate feedback and low constraint violation rate, we could learn experience from both extremely good and bad samples in replay buffer, which also contributes to the sampler's in-time adjustment in face of the reward-constraint imbalance. Specifically,
 define $\hat{r}$ as the average surrogate feedback of all samples in replay buffer, and $\sigma_r$ as the corresponding standard deviation with regard to the surrogate feedback. Then we collect $\frac{1}{6}$ batch of samples sampled  uniformly at random from samples in the buffer with surrogate feedback located outside $[\hat{r}-2\sigma_r,\hat{r}+2\sigma_r]$, so as
 to update the primal-dual Wolpertinger sampler.
 The same goes for the average constraint violation rate and the corresponding standard deviation.
 
(iii) The remaining undetermined samples are randomly selected from all samples in replay buffer.
 

 \subsubsection{G2ANet sampler}
 
 Although the primal-dual Wolpertinger sampler
 is capable of handling general forms of rewards and constraints, it does not leverage the characteristics of diversity constraints and  may not obtain nearly optimal solutions in all problem instances with such constraints. Therefore, to well infer the existence of diversity constraints  between every two arms, we ingeniously employ a graph attention network   named Game Abstraction based on Two-stage Attention Network (G2ANet) \cite{liu2020multi} as our G2ANet sampler. In the G2ANet sampler,
  each arm is regarded as an agent, as well as a node of  G2ANet.  The sampler determines whether there is an interaction between every two agents  and if the interaction exists, to estimate the importance of the interaction's influence on each agent's strategy, which can be utilized to infer the relationship between arms. We adapt G2ANet by applying our improved gumbel top-$K$ sampling technique  \cite{kool2019stochastic} to  obtain a binary elite sample from the outputs of  G2ANet in a differentiable manner and calculate the sampling probability in the loss term when performing back-propagation. 
  
  \textbf{Improved gumbel top-$K$ sampling.} \ 
At each round, G2ANet outputs a real number $out_i\in [0,1]$ for each node $i\in [L]$.  Based on
$\{out_i\}_{i\in [L]}$, we  adapt the    differentiable gumbel top-$K$ sampling algorithm (GTKS)
\cite{kool2019stochastic} to obtain the binary elite sample  $\{d_1,\cdots,d_L\}$ and estimate the sampling probability.

The original GTKS first 
performs the same data perturbation process as the gumbel-max sampling \cite{kool2019stochastic} on $out_i$ to obtain $out_i^{\prime}$ for $i\in [L]$. Then, draw an ordered sample of size $K$ as the top $K$ arms by $ I_{1}^{*}, \ldots, I_{K}^{*}=\arg \operatorname{top}_{K} \{out^{\prime}_{1},\cdots,out^{\prime}_{i},\cdots, out^{\prime}_{L}\} $ with $ out^{\prime}_{I_{1}^{*}} \geq out^{\prime}_{I_{2}^{*}} \geq \ldots,out^{\prime}_{I_{K}^{*}}$.  And
 calculate the ordered sampling probability $P(I_{1}^{*}\!=\!i_{1}^{*}, \ldots, I_{K}^{*}\!=\!i_{K}^{*})\!$.

However, here comes the problem that the choice of arms in this paper should be considered regardless of order during the top-$K$ sampling process, that is, the order of $out^{\prime}_{I_{1}^{*}} \geq out^{\prime}_{I_{2}^{*}} \geq \ldots \geq out^{\prime}_{I_{K}^{*}}$   should not be required.
  To  eliminate the above requirement, one should  perform $K!$ permutation for $I_{1}^{*}, \ldots, I_{K}^{*}$ and average the $K!$ ordered probabilities  to obtain the unordered sampling probability.
 To reduce the amount of calculation,  we only randomly generate $M$ permutations of  $I_{1}^{*}, I_{2}^{*}, \cdots, I_{K}^{*}$, and average the corresponding $M$ probabilities as the unordered sampling probability.

\textbf{Feedback after outputting a sample.} \
The composite feedback at each round   is calculated by $Surrogate(\cdot) -\lambda c(\cdot)$.
\subsubsection{Improved CEM-PPO evolutionary sampler}


Due to the considerable parameters and time-consuming 
back-propagation of deep neural networks, we seek for other methods that can solve  general online COPs in a parallel manner with fewer parameters and require less back-propagation, and cross-entropy method (CEM), a practical evolution strategy, just belongs to  such methods.  Basically, CEM is a search method based on parameter perturbation. It first imposes some reasonable perturbation on the parameter space $v$, and searches and selects some elite samples in the perturbed offsprings. Then the cross-entropy is utilized to guide the update of $v$, encouraging the perturbation to move towards the targeted optimization direction.
CEM is easy to be implemented and has strong universality in applications, and we slightly modify it to adapt to our setting. The detailed implementation is as follows:

\begin{itemize}
    \item \textbf{Initialization:}
    $\forall t\in [T], i\in [L]$, $A_{t,i}=1$ represents that arm $i$ is selected at round $t$, otherwise $A_{t,i}=0$. Assume that $A_{t,i}$ follows the Bernoulli distribution with  expectation $\mu_i$.    $\forall  i\in [L]$, we set $\mu_i=K/L$ and  $\bm{\mu}=\{\mu_1,\cdots,\mu_L\}$.
    
    \item\!\! CEM divides the total $T$ rounds into several epochs with length $N$.
    \begin{itemize}
        \item 
    For $Q\in \{0,1,2,\cdots, \lfloor  T/N  \rfloor -1\}, t\in \{QN+1,QN+1,\cdots,(Q+1)N\}$, sample $A_{t,i}$ from $Bernoulli(\mu_i)$. 
     \item 
    For $t=(Q+1)N$, calculate the composite score for each vector in $\{A_{QN+1},A_{QN+1},\cdots,A_{(Q+1)N}\}$ according to the elite sample evaluator in the master model.
     \item 
    Assume that $\{B_1,B_2,\cdots,B_{ \lceil \rho N  \rceil}\}$ are the $ \lceil \rho N  \rceil$ samples that rank the first $100\rho$ percentage on the composite score. Update $\bm{\mu}$ by $\bm{\mu}=\sum_{i=1}^{ \lceil \rho N  \rceil} B_i/ \lceil \rho N  \rceil$.   
\end{itemize}
  \end{itemize}
  
  To ensure that $\bm{\mu}$ can approximately portray the properties of the optimal solution region, we need to collect enough samples, i.e., $N$ should be large enough. However, 
    to maximize the accumulated composite score, the parameter $\bm{\mu}$ should be timely updated; otherwise, the number of bad samples would be largely due to delayed parameter updates and too much regret would be accumulated. 
  
  To address the above problem, \cite{gao2018post} divided each epoch into many small  intervals with length $n$. At each intersection of two intervals, \cite{gao2018post} applied its Joint Learning with Proximal Policy Optimization (\textit{Post}) technique to keep the parameter $\bm{\mu}$ updated more frequently. For $j \in\{0,1,2, \ldots,  \lfloor N/n   \rfloor -1\}$, when $t=Q N+(j+1) n$,
  \textit{Post} customizes an objective of proximal policy optimization  as follows:
 \begin{equation}
      \begin{aligned}
     J_t=& \max _{u_{new}} \frac{1}{n} \sum_{t=Q N+j n+1}^{Q N+(j+1) n} \sum_{i=1}^{L} \frac{P(A_{t, i} \mid u_{n e w, i})}{P(A_{t, i} \mid u_{o l d, i})}(Score(A_{t})
      \\&-b) -\beta D_{K L}(P(A_{t, i} \mid u_{o l d, i}) \| P(A_{t, i} \mid u_{n e w, i})),
  \end{aligned}
 \end{equation}
  where $u_{old}$ is the parameter vector the CEM sampler follows at rounds $\{QN+jn+1, QN+jn+2,\cdots,QN+(j+1)n\}$, and $u_{new}$ is the parameter vector the CEM sampler follows at rounds $\{QN+(j+1)n+1,QN+(j+1)n+2,\cdots,QN+(j+2)n\}$. If $A_{t,i}=1$, set $\mathbb{P}(A_{t, i} \mid u_{i})=u_{i}$. If $A_{t,i}=0$, set $\mathbb{P}(A_{t, i} \mid u_{i})=1-u_{i}$.
  
  Every $n$ rounds, update $J_t$ and implement several gradient descent steps on $J_t$  with regard to $u_{new}$. 

  As to our improvements,  
 1) to extract information from more elite samples with fewer sampling trials, we update the parameters of CEM-PPO partially ($\frac{1}{2}$) based on the best global elite samples provided by all samplers evaluated by the surrogate feedback in the latest epoch; 2) to keep the  update of  parameters more stable and avoid sticking into local optimal solutions,
  we  save the best elite samples in history and mix them with current elite samples when updating CEM's parameters; 3) also, a discounted decaying factor   is  utilized to combine current parameters with previous parameters; 4) furthermore,
  we apply  our adapted gumbel top-$K$ sampling trick to sample binary elite samples from  $v$ and
  adjust the sampling probability term in the PPO-like objective, accordingly. Compared to the original Bernoulli sampling technique, our new sampling trick has the potential to model a more complicated relationship among arms.  
  
  \subsubsection{Random sampler}
  The random sampler  samples  actions in $\mathcal{A}_0$ uniformly at random for enough exploration, and also keeps the elite sample with the best $Score$ in history from all samplers.
  
\subsubsection{Teacher-student sampler}
Since we collect a set of elite samples from multiple samplers, it is natural to allow samples to learn from each other, rather than see each sample as an independent one. Inspired by this idea, 
we take Teaching Learning Based Optimization (TLBO) \cite{crawford2020teaching} as our teacher-student sampler   to improve the overall quality and diversity of all global elite samples.
  TLBO  is a population method based on the knowledge
that a teacher has the potential to
improve the knowledge level of the class. Moreover,
students can also switch their knowledge for mutual assistance. Here we use TLBO to  perturb and improve the elite samples from the other slave samplers, avoiding local optimal solutions.
Denote $S_t$ as the student set which consists of elite samples from other slave models at round $t$. Take the sample  with the highest $Score$, calculated by the elite sample evaluator in the master model, as the teacher $T$. 
TLBO allows the teacher-student and student-student interactions at each round $t$:
  \textbf{(i) Teacher-student interaction:}
   Select a student sample $A$ from $S_t$ uniformly at random. Set $B=A+rand*(T-A)$ where $rand$ is a random variable obeying the uniform distribution in $[0,1]$. 
   Rank all components in $B$ and reform $B$ by setting $B_i=1$ if the original $B_i$ belongs to the largest $K$ components in the original $B$; otherwise, set $B_i=0$. 
\textbf{(ii)  Student-student interaction:}
     Select two student samples $A$ and $B$ from $S_t$ uniformly at random. If $Score(A)<Score(B)$, set $C=A+r a n d *(B-A)$; otherwise, set $C=A+r a n d *(A-B)$.  Rank all components in $C$ and reform $C$ by setting $C_i=1$ if the original $C_i$ belongs to the largest $K$ components in the original $C$; otherwise, set $C_i=0$.

Now we obtain  new elite samples $B$ and $C$. Repeat the above procedures several times to generate more  elite samples.

   

%% file: 3_experiment.tex
\section{Experiments}
To demonstrate the efficiency of our  master-slave model, we apply it to solve top-$K$ recommendation problems with non-linear bandit feedback under diversity constraints over four synthetic datasets and four real-world datasets, and compare it against several existing works, including Reinforcement Learning from Demonstrations (RLfD) \cite{gong2019exact},  Wolpertinger architecture \cite{dulac2015deep},  Primal-dual Wolpertinger architecture (PDWA) \cite{dulac2015deep,tessler2018reward}, Primal-dual G2ANet (PDG2ANet) \cite{tessler2018reward,liu2020multi}, Improved CEM-PPO evolutionary algorithm  \cite{gao2018post},  Gumbel top-$K$ reinforce (GTKR) \cite{kool2019stochastic}, CSAR-EST1, CSAR-EST2 \cite{rejwan2020top}, CascadeKL-UCB \cite{kveton2015cascading}, CascadeLSB \cite{hiranandani2020cascading}, and CascadeHybrid \cite{li2020cascading}. 
Besides, we also calculate the ground truth by Gurobi. Most of the experimental results are placed in the \textbf{Supplementary Material} file due to space limitations.



\subsection{Evaluation metric}
Define the average recommended rate of slave model $\mathcal{S}$  as the average percentage that the elite samples provided by $\mathcal{S}$ are truly recommended by the master   from round $2L+1$ (the time when all  slave models begin to work) to round $t$.
In experiments,
we adopt three evaluation metrics to measure the performance of  algorithms and the importance of each slave model, including (i) $r_t$, the reward feedback which indicates the user's preference for recommended items; (ii) $c_t$, the constraint violation rate which shows algorithms' ability to deal with diversity constraints; and (iii) $arr_t$, the average recommended rate which infers the importance of a slave model  in providing useful samples for the master model.
\subsection{Experiments on synthetic datasets}

In the synthetic dataset, we set $L=300$, $f_{in}=20$, $K=20$, $\lambda=5$, $\epsilon_0=0.05$, $\rho=0.1$, and $d=10$, where $d$ is the dimension of each item's feature vector. The feature vectors are generated uniformly at random from $[0,1]^d$, and by setting the normalized edit distance (NED)  threshold $\tau$  to be $0.5$, we obtain $3962$ diversity constraints.   Let $r_t=h(A_t)+\epsilon_t$  with $\epsilon_t$ chosen from the normal distribution with mean $0$ and standard error $0.1$. And we
set $h(A_t)$ at round $t$  to be $\theta^{\top} A_t$, $(\theta^{\top} A_t)^3$, $A_t^{\top} Q A_t$ and $(\theta^{\top} A_t)^2+A_t^{\top} Q A_t$ respectively in different  experiments, where $\theta$ is a vector randomly selected from $[0,0.5]^L$ and $Q$ is a matrix randomly selected from $[0,0.5]^{L\times L}$.
All algorithms are run over $T = 5k$  rounds for $10$ times  
with results shown in Figure \ref{fig:Synthetic dataset}.

For $r_t$ displayed in Figure \ref{fig:Synthetic dataset}, our master-slave algorithm is 8.24\%, 6.38\%, 17.88\%, and 9.11\% better than RLfD, the second-best algorithm in   four synthetic datasets, respectively, significantly outweighing the performance of each individual slave algorithm due to the novel master-slave interactions and policy co-training technique.
For $c_t$ in Table \ref{tab:constraint}, our  algorithm  is 59.09\%, 55.17\%, 76.08\%, and 62.5\% better than  the second-best algorithm in each   dataset, which shows the powerful ability to deal with constraint satisfaction of our master-slave architecture.
As to $arr_t$ shown in  Figure \ref{fig:Recommended Rate syn},
the  random sampler with  keeping the best elite sample in history (RS) and the improved CEM-PPO sampler (I-CEM-PPO) are the leading slave models to provide the most elite samples, followed by the solver sampler (SS).
In contrast, the others work out mainly at an early stage which provide potentially great  samples to pave the way for the great performance of RS, I-CEM-PPO, and SS at a later stage.

\subsection{Experiments on real-world datasets}
We extract four real-world datasets from four recommendation platforms named LastFM,  Delicious,   MovieLens\footnote{LastFM, Delicious, MovieLens: https://grouplens.org/datasets
/hetrec-2011/} and   Today Module on Yahoo! Front Page\footnote{https://webscope.sandbox.yahoo.com/catalog.php?datatype=r\&did=49}, respectively. 
All the above datasets contain user-item-tag logs which record the 
   behavior that a user tags an item at a certain moment and assigns the corresponding tag id.  Table \ref{tab:realdata} provides the detailed statistics of the LastFM, Delicious, MovieLens, and Yahoo! datasets. 
Note that the full Yahoo! dataset contains several subsets, here we only include the first one.
To further utilize its additionally provided static features for users and items, we treat users/items with the same feature vector up to a one-decimal rounding as the same user/item, resulting in $290$ users and $20$ items. 
And therefore we do not need to further cluster items to control the item set size in this dataset.

With regard to extracting  feature vectors of items, we directly borrow items' feature vectors in Yahoo! since it is the only dataset that provides feature vectors. As for the LastFM, Delicious, and MovieLens datasets, we mostly follow the data preprocessing procedures in \cite{cesa2013gang}   to construct feature vectors of items.   Specifically, we first compute the tf-idf representations of the tags and extract the first $10$ principal components as the tag vectors. Then we make a pass over all logged events. For each user-item-tag tuple, we go through
the original dataset and add each tag vector to the corresponding item vector (which is initialized as a zero vector).
After the pass, all item vectors are normalized to unit vectors. To control the size of the item set, we group all items into $40$ clusters by K-means and use the mean vector of each cluster as an item feature.
The data log of the most active user is adopted.
 
 For the reward feedback, we adopt real clicks of the most active user in the original datasets with results shown in Figure \ref{fig:Real-world datasets}\footnote{Since we strictly follow the data log, the algorithms are run for just one time.}. 
Specifically, define $Card_{real,t}$ as the item set that the user clicks during the $2K(K=10)$ rounds closest to time $t$, and $Card_{algo,t}$ as  the item set that an algorithm decides to 
recommend at time $t$.  Then the reward  at time $t$ is set to be $\frac{|Card_{real,t}\cap Card_{algo,t}|}{2K}$. Apart from adopting the real feedback based on real user clicks, we also construct four synthetic feedbacks based on items' information
 for Yahoo! and Movielens datasets.
The results are shown in Figure \ref{fig:yahoo dataset} and \ref{fig:MovieLens dataset}. And the procedure of feedback construction is elaborated as follows:

     \textbf{(i) Yahoo! dataset:} 
 Calculate the click rate vector for the most active user where its $i$-th component represents the number that the user clicks on the $i$-th item. Normalize this vector to get $\theta$ and set the expected  user feedback $h(A_t)$ at round $t$ to be $\theta^T A_t$, $(\theta^T A_t)^3$, $A_t^T Q A_t$, and $(\theta^T A_t)^2+A_t^T Q A_t$, respectively, in different  experiments, where $Q$ is a matrix randomly selected from $\mathbb{R}^{L\times L}$. 
 
 \textbf{(ii) Movielens dataset:}
(a) To increase the training difficulty, we re-cluster the item vectors into $100$ groups and set $L=100$,  $f_{in}=20$, $K=10$, $\epsilon_0=0.05$, and $d=10$.  By setting the normalized edit distance (NED)  threshold $\tau$  to be $1$, we obtain $730$ diversity constraints.
(b)
Set $\theta$ to be the feature vector of the most active user and $h(A_t)$ is set in the same manner as that for the Yahoo! dataset.  

As to  $r_t$,  our master-slave algorithm is 40.10\%, 56.73\%, 18.87\%, and 27.27\% better than Wolpertinger, the second-best algorithm,  in the  four real datasets with real feedback, showing the excellent performance of our algorithm in non-stationary environments.
In  Yahoo! and MovieLens  with  synthetic feedback, our algorithm also achieves the best performance in all settings. 
As to $c_t$ shown in Table \ref{tab:constraint}, our master-slave algorithm  is 73.91\%, 80.00\%, 69.23\%, and 69.70\% better than the second-best algorithm in each  real  dataset.
As to the average recommended rate  shown in Figure \ref{fig:Recommended_Rate_real}, 
the solver sampler and the random sampler with keeping the best elite sample in history occupy more and more dominant positions in providing useful elite samples at a later stage, while the others work out well  at the beginning but gradually 
give less and less important advice to our master model. The relatively balanced importance of slave models across different stages displays our reasonable selection and configuration of those slave models. 

%% file: 4_conclusion.tex
\section{Conclusion}
In this paper, we study a top-$K$ combinatorial multi-armed bandits problem under non-linear rewards and diversity constraints and propose a master-slave hierarchical architecture to efficiently explore the combinatorial and constrained action space.  
Six slave models with distinguishing advantages are  designed to provide diversified and elite samples for the master model, and an improved Gumbel top-$K$ sampling trick is adopted to ensure the end-to-end differentiable pipeline. Thanks to the elaborate  design of multiple slave models, interactions between the master and slave models, and policy co-training mechanism, our method achieves remarkable performance  in  extensive experiments on both synthetic and real datasets.
In the future, we  will try to
 adopt top-$K$ off-policy correction  \cite{chen2019top} and adversarial training  for deeper co-training  of  slave models.

%% file: 5_appendix.tex
\newpage
\section{Appendix}

In the Appendix, we first display the   experimental details that we miss in the main text in Section \ref{figtab}. Then we describe the details of  our master model and multiple slave models in Section \ref{master}
and Section \ref{slave}. Next in the discussion part, we systematically sort out the novelty of our method in Section \ref{novelty}, clarify some key points that might confuse readers in Section \ref{clarify}, and deliberately adapt the algorithm for generalization to  non-stationary environments (Section \ref{nonstationary})  and to other combinatorial optimization problems (Section \ref{othercop}).

\subsection{Experimental details}\label{figtab}
\subsubsection{Brief introductions of baselines}
\begin{itemize}
\item \textbf{Reinforcement Learning from Demonstrations (RLfD) \cite{gong2019exact}:}
\cite{gong2019exact} proposed  RLfD to deal with exact-k recommendation via maximal clique optimization.
We choose its most competitive "RL(w/hill-climbing) + SL(w/policy-sampling)"  mode to compare with our algorithm\footnote{https://github.com/pangolulu/exact-k-recommendation} and use $r_t$ as the feedback at time $t$.
    \item \textbf{Primal-dual Wolpertinger architecture (PDWA):}
    This is adapted from our primal-dual Wolpertinger sampler in the slave models which integrates  Reward Constrained Policy Optimization with the Wolpertinger architecture used to  handle large discrete action space\footnote{https://github.com/ChangyWen/Wolpertinger\_ddpg, https://github.com/xkianteb/ApproPO/blob/master/ApproPO/rcpo.py}. 
    The only difference is that we apply $r_t-\lambda_2 c_t$ as
    the composite feedback 
    instead of $Surrogate(.)-\lambda_2 c(.),$ where 
    $Surrogate(.)$ refers to the $U$ function in (\ref{eq:U}).
    \item \textbf{Primal-dual G2ANet:}
    This is adapted from the Primal-dual Wolpertinger architecture\footnote{https://github.com/starry-sky6688/StarCraft/blob/master/network/G2ANet.py}. The only differences are 
     
        \textbf{(i)} we replace the MLP actor network  in the  Wolpertinger Architecture with  G2ANet;
        
        \textbf{(ii)} we convert the proto-action to a real recommendation action in the same manner as the G2ANet sampler, instead of the $K_1$-nearest manner;
       
        \textbf{(iii)} we apply $r_t-\lambda_2 c_t$ as
    the composite feedback.
    \item \textbf{Improved CEM-PPO evolutionary algorithm:}
    This is adapted from our improved CEM-PPO evolutionary sampler. The only difference is that we apply $r_t-\lambda c_t$ as
    the composite feedback 
    instead of $Surrogate(.)-\lambda c(.)$.
    \item \textbf{Gumbel top-$K$ reinforce (GTKR):}
 This is adapted from the Reinforce RL algorithm. The only difference is that we utilize the improved gumbel top-$K$ sampling technique so as to transform the network's real-valued outputs  into a binary vector for recommendation\footnote{We adapt the code in https://github.com/chingyaoc/pytorch-REINFORCE}.  We use $r_t$ as the feedback at round $t$.
    
    \item \textbf{CSAR-EST1 and CSAR-EST2:}
 \cite{rejwan2020top}    proposed Combinatorial Successive Accepts and Rejects (CSAR) based on a sampling method utilizing Hadamard matrices to estimate arms with a small number of samples. CSAR-EST1 and CSAR-EST2 are two representations of CSAR with different  estimation tricks that estimate the expected rewards for all the arms.
    \item \textbf{CascadeKL-UCB:}
 \cite{kveton2015cascading}  formulated  the
  cascading bandits setting as a stochastic combinatorial
partial monitoring problem and introduced CascadeKL-UCB, a KL-UCB based algorithm, to deal with this partial monitoring problem.

    \item \textbf{CascadeLSB:}
\cite{hiranandani2020cascading} proposed CascadeLSB, a  click model which considers both position bias and diversified retrieval in cascading bandits.
    
    \item \textbf{CascadeHybrid:}
 \cite{li2020cascading} formulated the setting of  online learning to rank for relevance and diversity as a cascade hybrid bandits problem and introduced the CascadeHybrid algorithm that utilizes a hybrid model to deal with this problem.
    
    \item \textbf{Ground truth:}
 For experiments with stationary and synthetic user feedback, we calculate the optimal reward when all environmental information is known and all the diversity and cardinality constraints are satisfied by the Gurobi solver and regard it as the ground truth. Therefore, ground truth is the solution that satisfies all constraints but may not own the highest reward globally.
\end{itemize}

 \subsubsection{Influence of the reward-constraint trade-off coefficient $\lambda$}
 We vary  $\lambda$ from $0.1$ to $50$ to see the performance of each baseline under different 
    reward-constraint trade-off coefficients. As can be seen in Table \ref{tab:syn_lambda} and Table \ref{tab:del_lambda}, our master-slave algorithm outperforms other baselines nearly all the time however $\lambda$ changes.
    
\subsubsection{Analysis of each slave sampler}
 We test the performance of each single slave sampler (by adopting the master-slave architecture with only one slave sampler) except the teacher-student sampler, which is not capable to be implemented individually, in both synthetic and real-world datasets. As can be seen in  Table 
 \ref{tab:single}, the solver sampler achieves the overall  best performance and the random sampler with keeping the best sample in history performs worst in total. The G2anet sampler, the CEM-PPO sampler, and the primal-dual wolpertinger sampler have similar and medium performance.

Besides, we also record the average recommended rate of each slave sampler when performing the   master-slave algorithm with all slave samplers in both synthetic and real-world datasets. As can be seen in Figure \ref{fig:Recommended Rate syn} and Figure \ref{fig:Recommended_Rate_real}, the solver sampler and the 
random sampler with keeping the best sample in history perform the best overall. The CEM-PPO achieves a great average recommended rate sometimes in synthetic datasets but performs relatively normally in real-world datasets.  The others make a relative big difference at the beginning and gradually fall behind the solver sampler and the random sampler with keeping the best sample in history later on. To sum up, each sampler provides a  non-neglectable contribution to our master-slave architecture.

\subsubsection{Ablation study}
\begin{figure*}[ht]  

\centering
\includegraphics[width=0.24\linewidth]{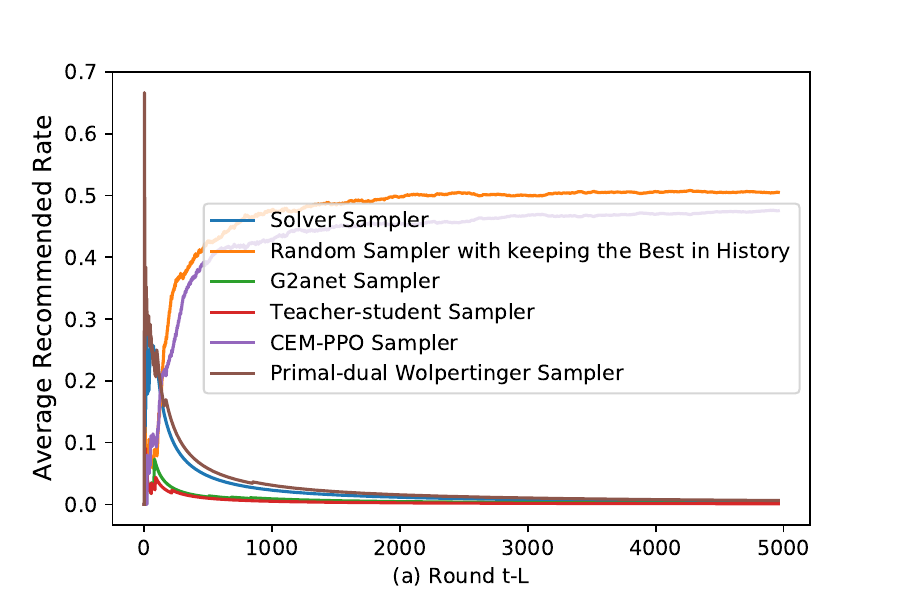} \!\!\! 
\includegraphics[width=0.24\linewidth]{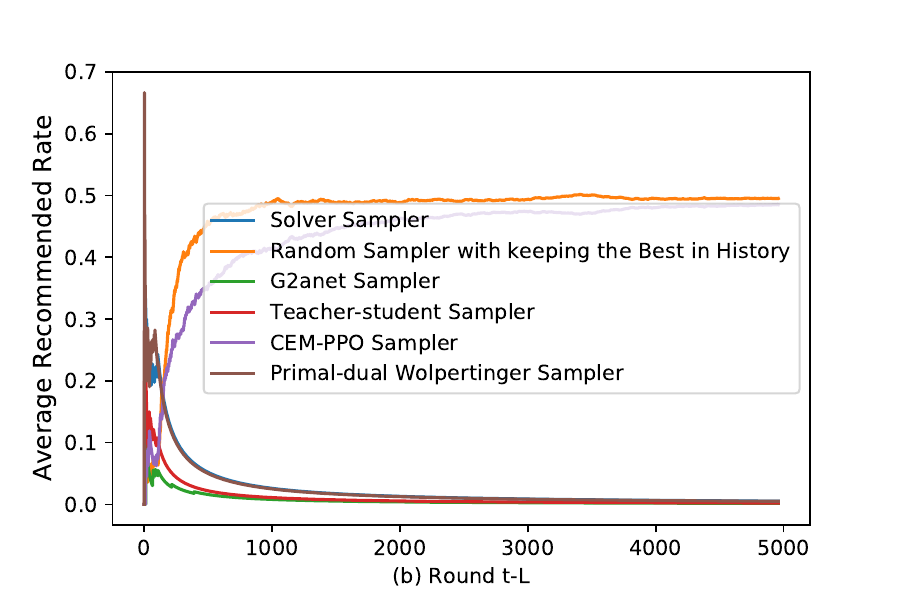} \!\!\!
\includegraphics[width=0.24\linewidth]{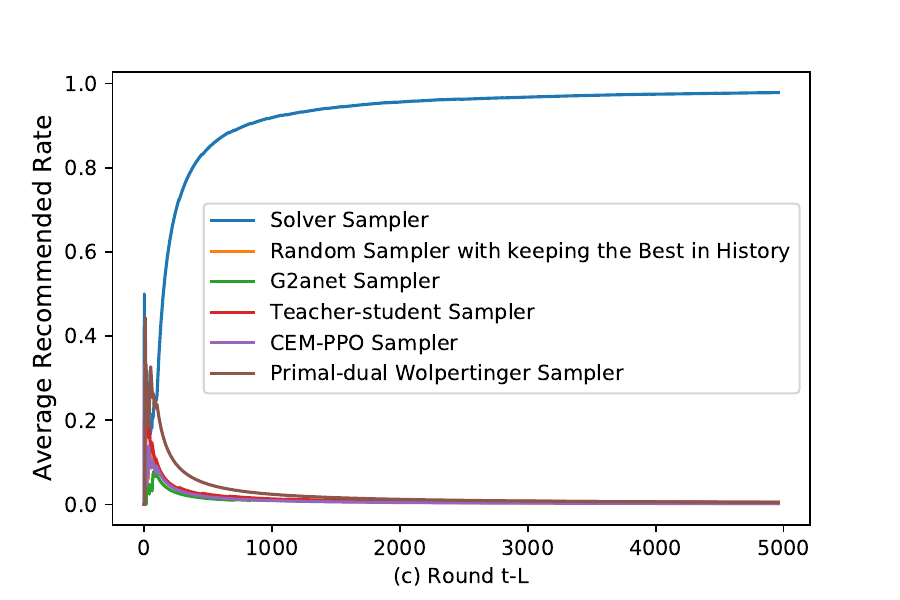} \!\!\!
\includegraphics[width=0.24\linewidth]{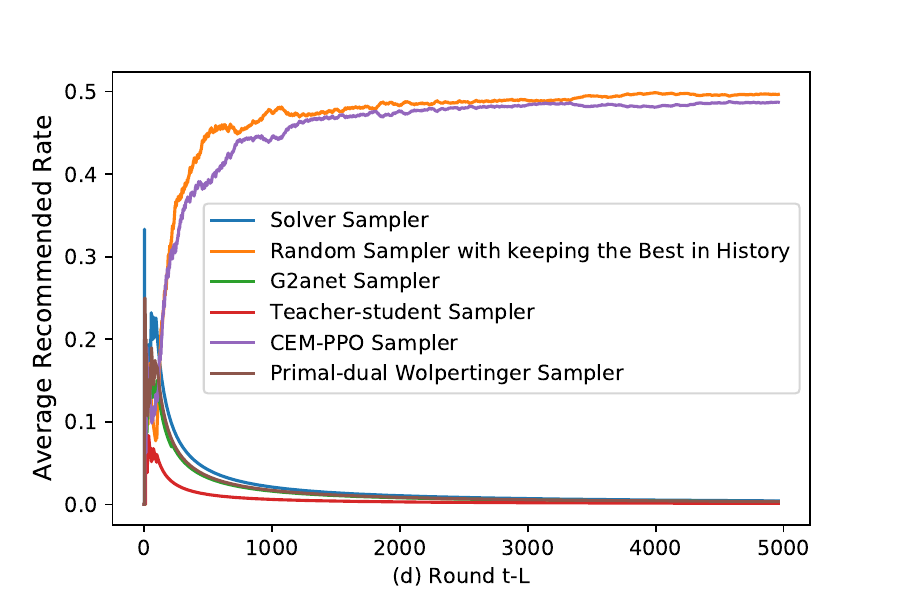} 
\caption{Average recommended rate of six slave models in  synthetic datasets: (a) Linear form; (b) Quadratic form; (c) Cubic form; (d) Mixed form.}
  \label{fig:Recommended Rate syn}
\end{figure*}

\begin{figure*}
 
\centering
\includegraphics[width=0.24\linewidth]{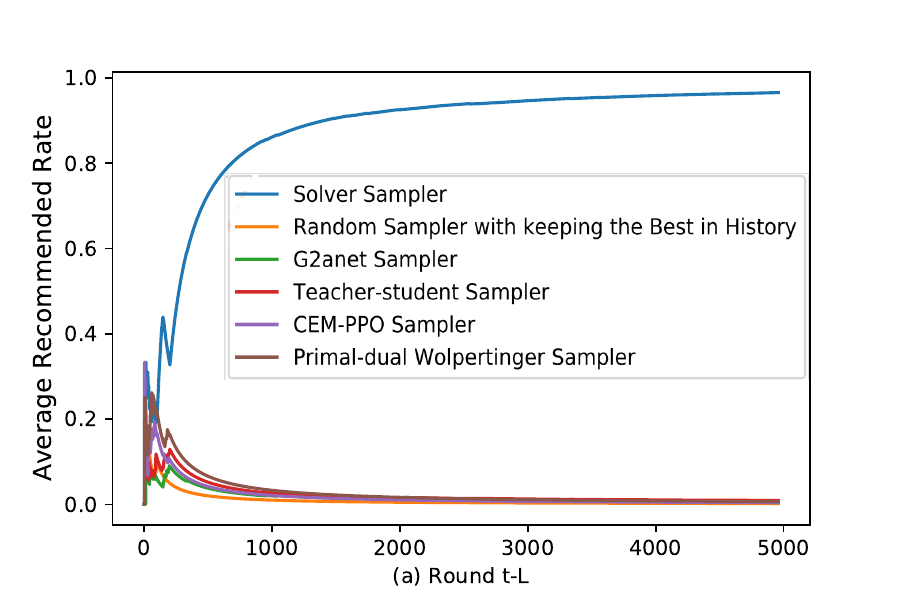} \!\!\!
\includegraphics[width=0.24\linewidth]{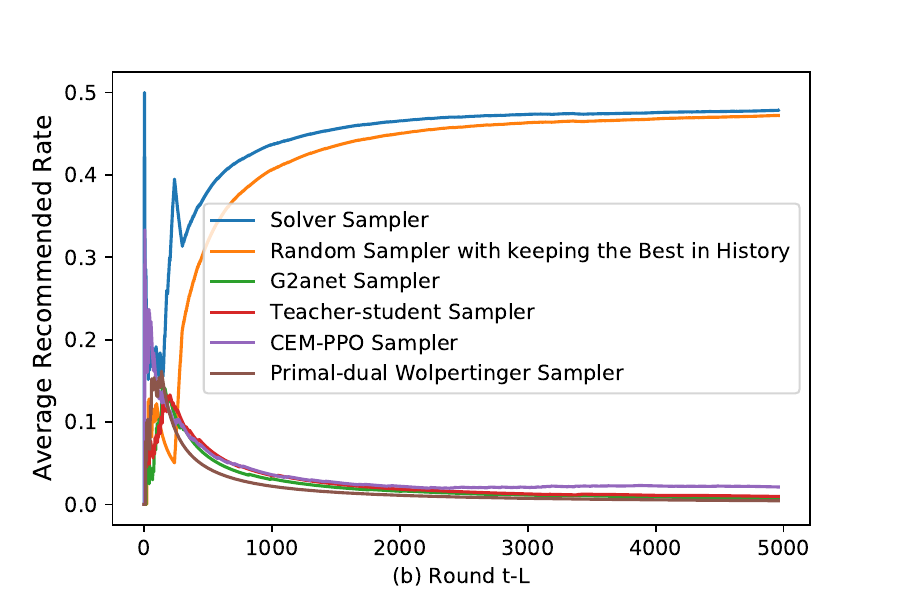} \!\!\!
\includegraphics[width=0.24\linewidth]{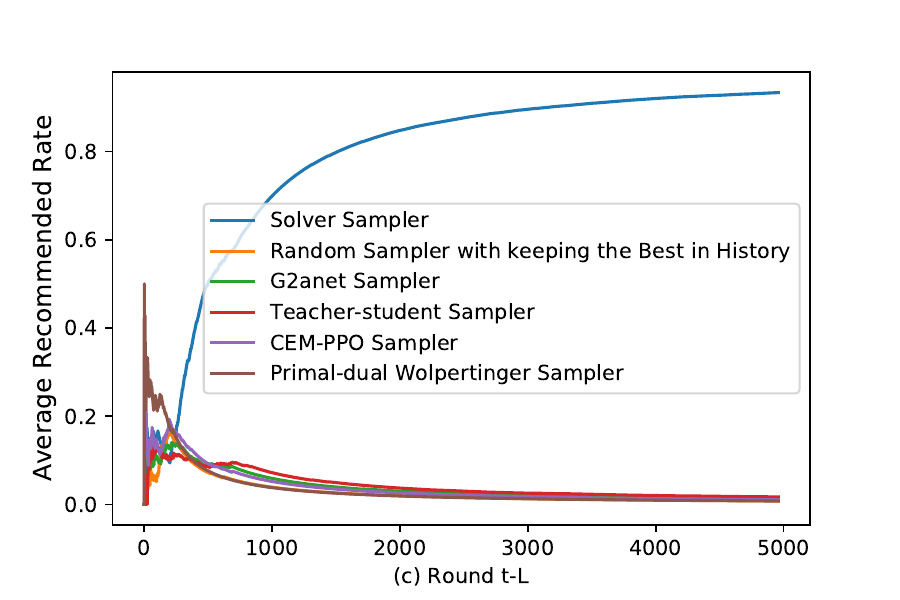} \!\!\!
\includegraphics[width=0.24\linewidth]{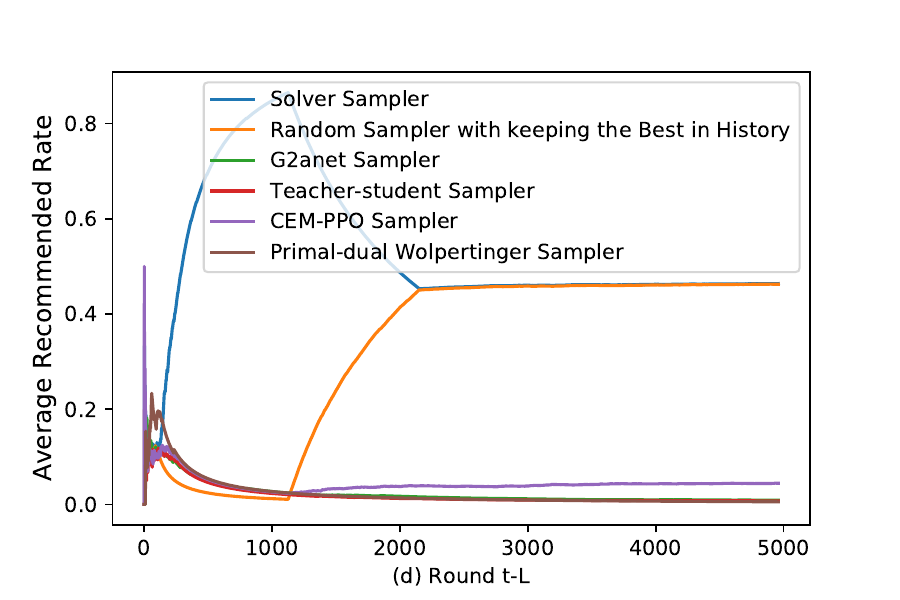} 
  \caption{Average recommended rate of six slave models in real-world datasets:  (a) Delicious; (b) LastFM; (c) MovieLens; (d) Yahoo!.}
  \label{fig:Recommended_Rate_real}
\end{figure*}

\begin{figure*}[th!] 
\centering
\includegraphics[width=0.24\linewidth]{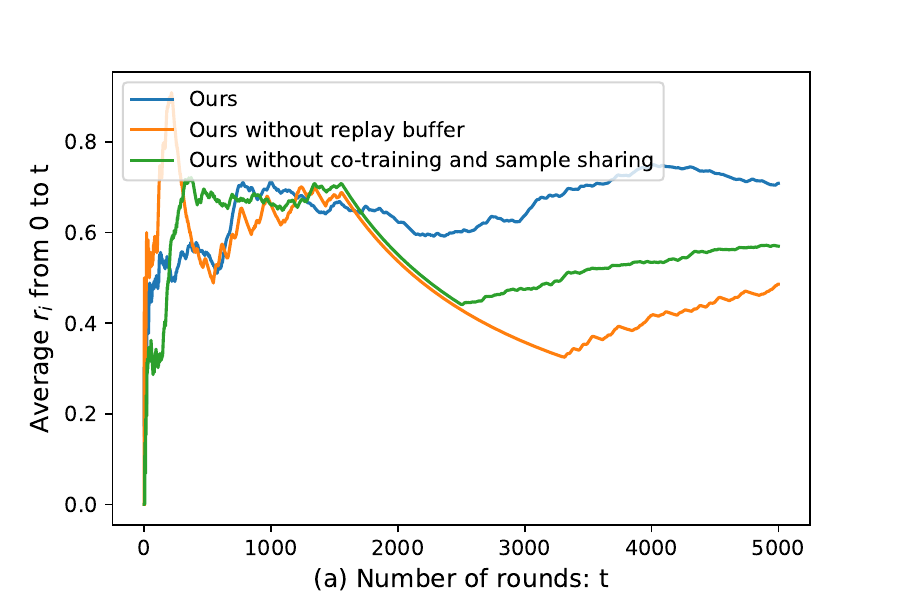} \!\!\!
\includegraphics[width=0.24\linewidth]{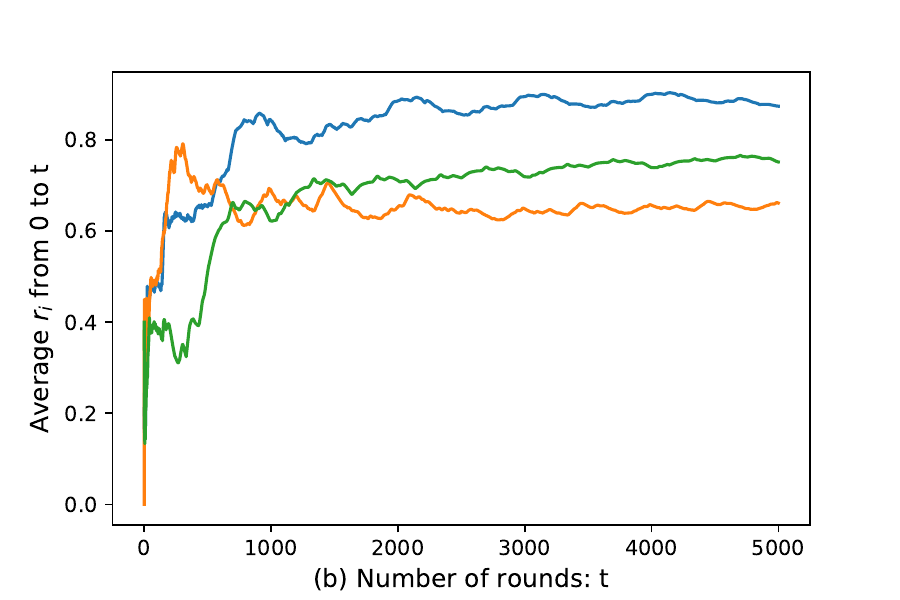} \!\!\!
\includegraphics[width=0.24\linewidth]{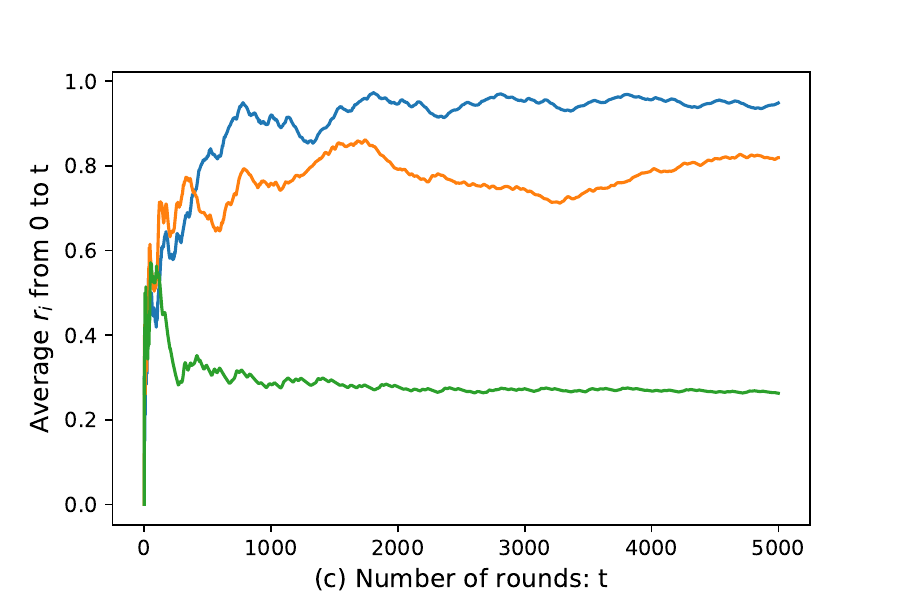} \!\!\!
\includegraphics[width=0.24\linewidth]{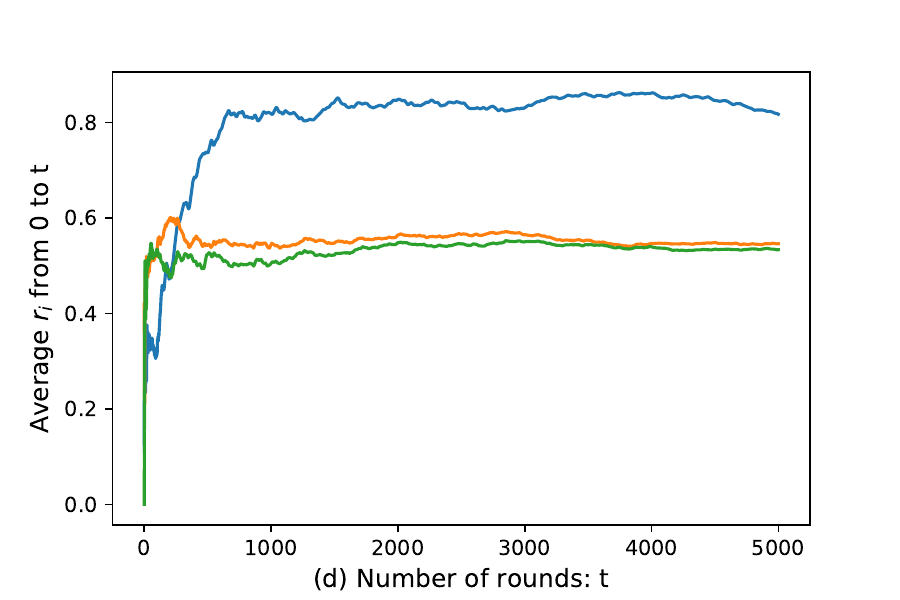} 
  \caption{Ablation study in real-world datasets:  (a) Delicious; (b) LastFM; (c) MovieLens; (d) Yahoo!.}
  \label{fig:ablation}
\end{figure*}

We perform ablation study on both synthetic and real-world datasets by
(i) comparing $arr_t$ of all  slave models (Figures \ref{fig:Recommended Rate syn}  and  \ref{fig:Recommended_Rate_real}).
(ii) comparing the performance of our method with and without each slave sampler (Table \ref{tab:ablation_study_reward}).
(iii) comparing the performance of our method with and without the prioritized experience replay buffer mechanism (Figure \ref{fig:ablation}). 
(iv) comparing the performance of our method with and without the policy co-training mechanism (Figure \ref{fig:ablation}).

From Figure \ref{fig:ablation}, both the customized prioritized experience replay buffer mechanism and the policy co-training technique  make a great difference in the performance improvement of our master-slave algorithm. Moreover, from Table \ref{tab:ablation_study_reward},
the lack of any slave sampler brings varying degrees of reduction in algorithm performance, which consistently demonstrates the precision of algorithm design and the importance of tacit cooperation among different slave samplers. 
 
 \subsubsection{Hyper-parameter analysis}
\begin{figure*}[th!] 
\centering 
\includegraphics[width=0.19\linewidth]{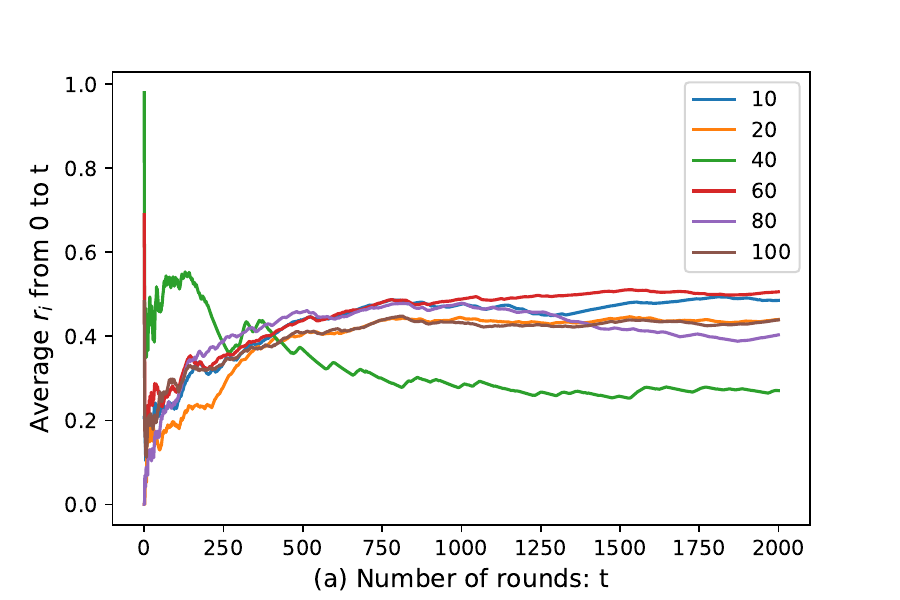} \!\!\!
\includegraphics[width=0.19\linewidth]{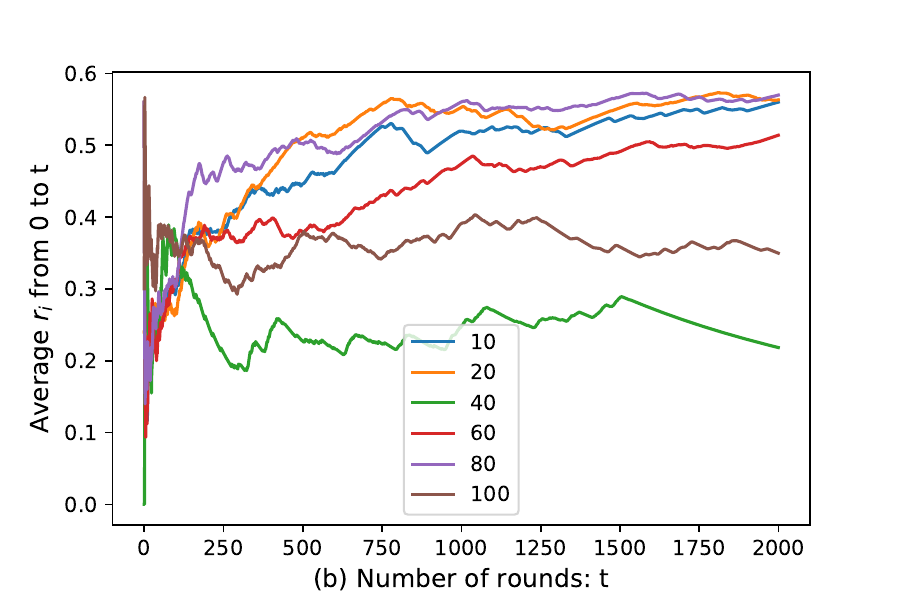} \!\!\!
\includegraphics[width=0.19\linewidth]{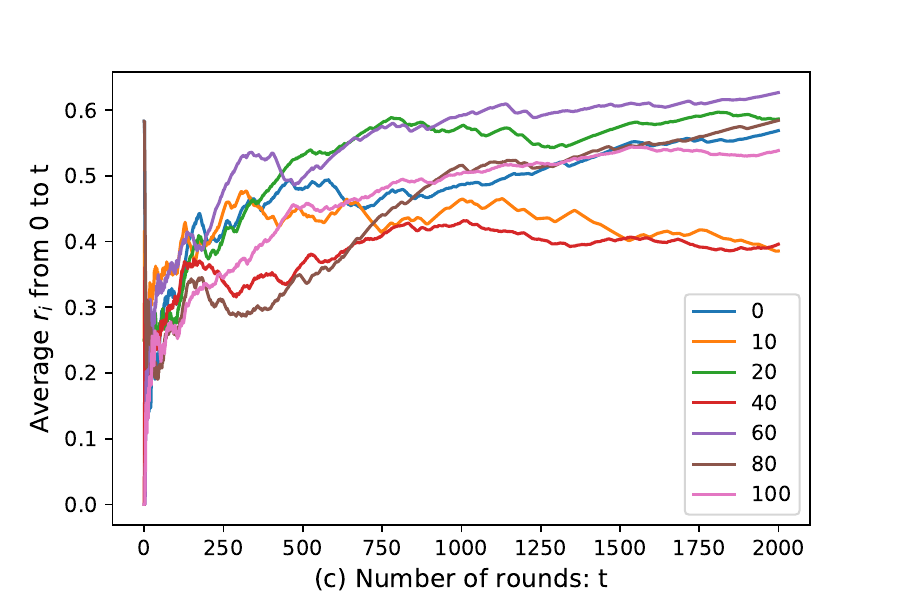} \!\!\!
\includegraphics[width=0.19\linewidth]{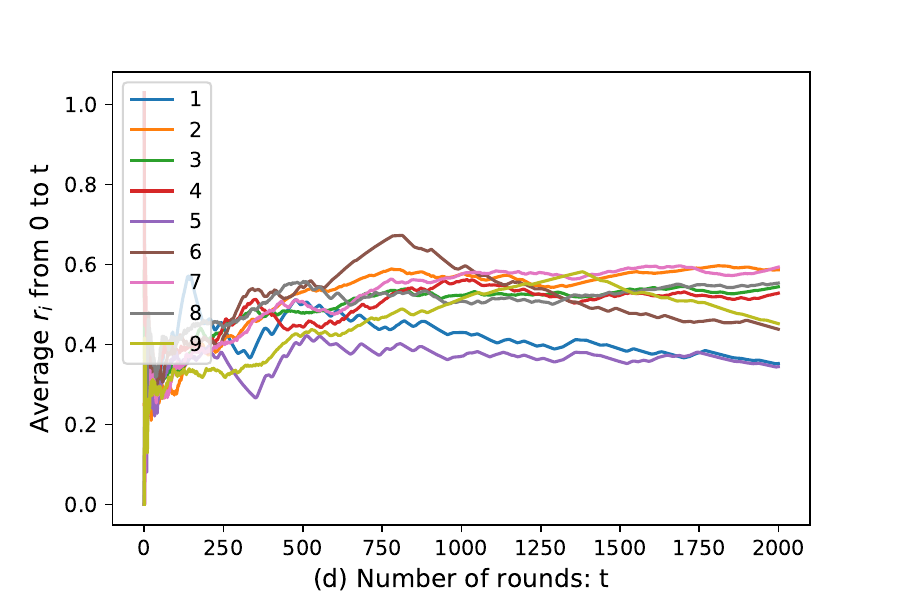} \!\!\!
\includegraphics[width=0.19\linewidth]{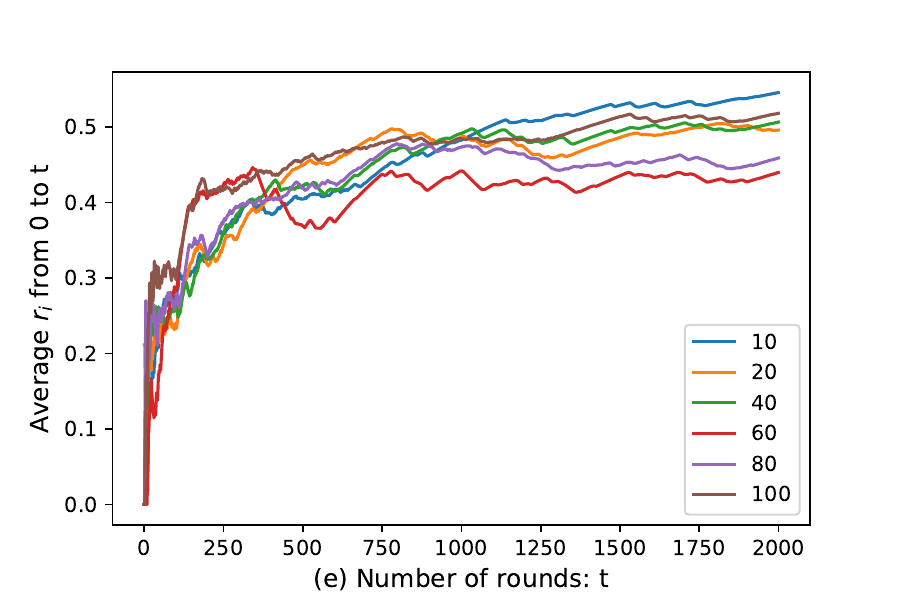} 
  \caption{Hyper-parameter analysis on the MovieLens dataset:  (a) $n_{es}$; (b) $n_{c}$; (c) $n_{D}$; (d) $n_{re}$; (e)  $f_{in}$.}
  \label{fig:hyper-ml}
\end{figure*}

\begin{figure*}[th!] 
\centering
\includegraphics[width=0.19\linewidth]{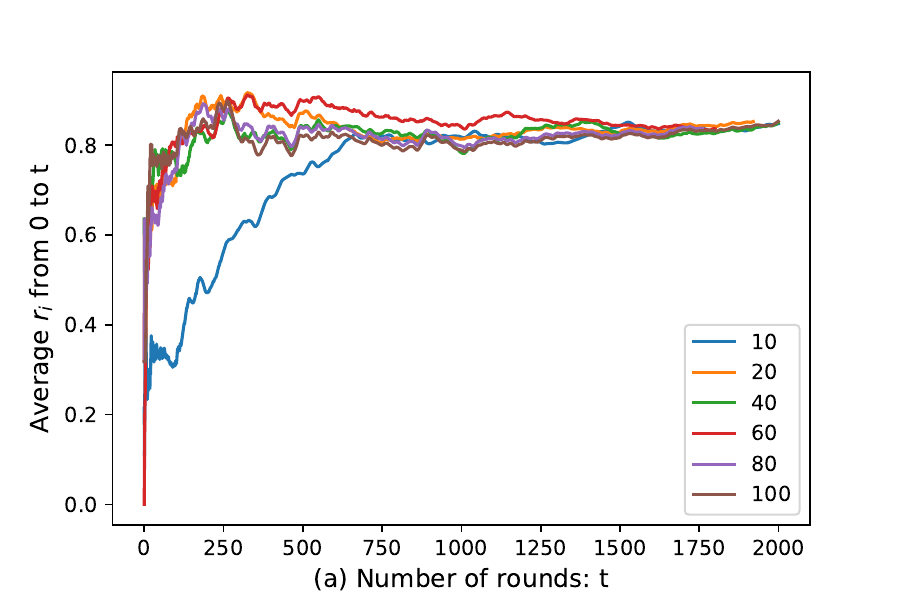} \!\!\!
\includegraphics[width=0.19\linewidth]{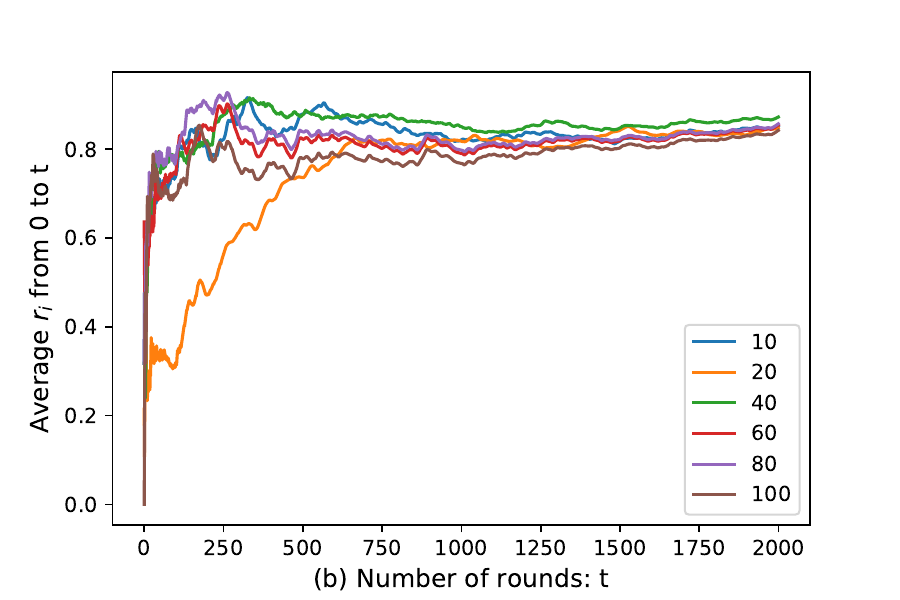} \!\!\!
\includegraphics[width=0.19\linewidth]{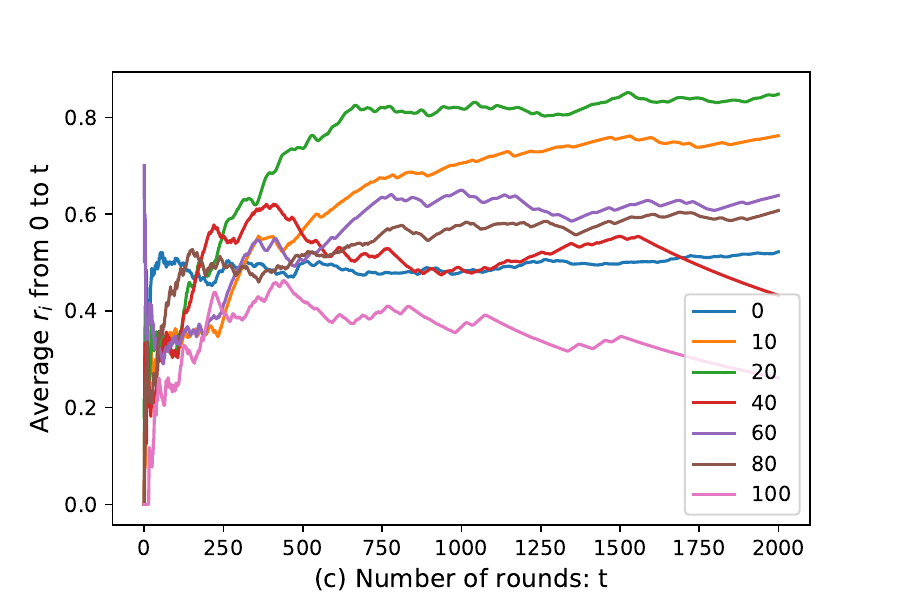} \!\!\!
\includegraphics[width=0.19\linewidth]{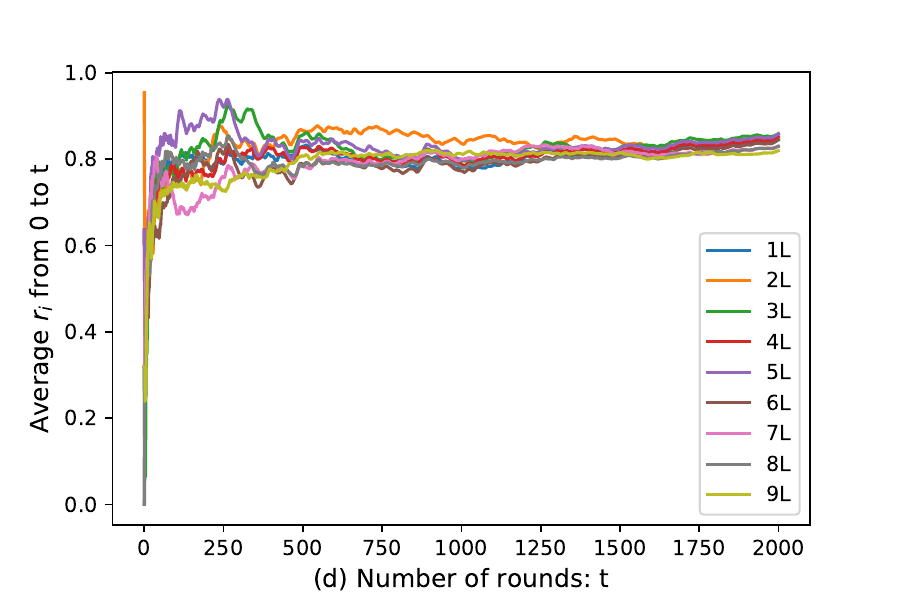} \!\!\!
\includegraphics[width=0.19\linewidth]{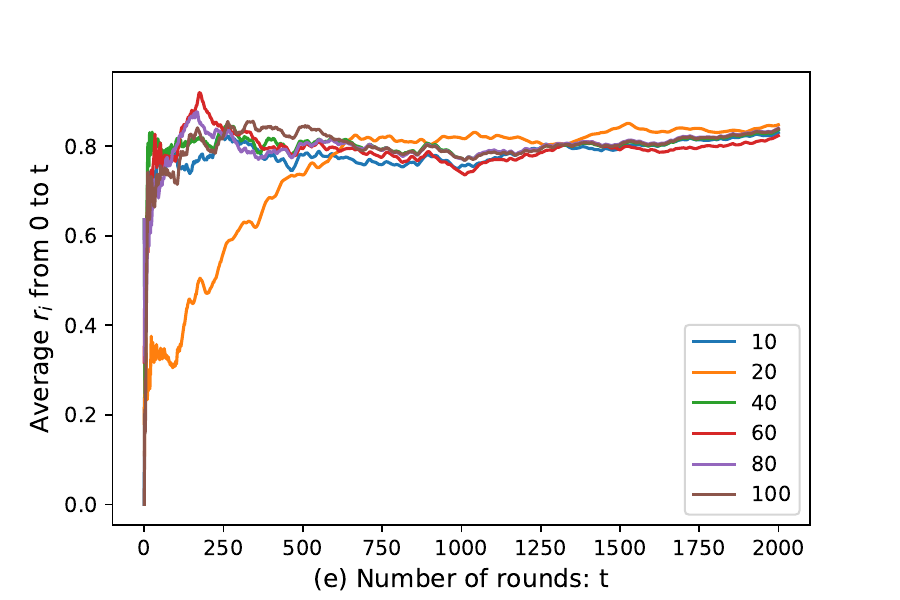} 
  \caption{Hyper-parameter analysis on the  Yahoo! dataset.}
  \label{fig:hyper-yahoo}
\end{figure*}

In this section, we perform the  hyper-parameter analysis for our algorithm on the MovieLens and the Yahoo datasets with results shown in Figures \ref{fig:hyper-ml} and \ref{fig:hyper-yahoo}. Denote $n_{es}$ as the number of total elite samples per round, $n_{c}$ as the number of clusters in replay buffer of the PDWA slave sampler, $n_{D}$ as iterations of learning from demonstrations each time,  $n_{re}$ as the number of rounds for random exploration, and $f_{in}$ as the interaction frequency between the master and the slave models. 
We first set $n_{es}=10,n_{c}=20,n_{D}=20,n_{re}=2L$, and $f_{in}=20$.
Then (i) fix $n_c,n_D,n_{re},f_{in}$, see the performance of our method when setting $n_{es}=10,20,40,60,80,100$, respectively (Figure \ref{fig:hyper-ml},\ref{fig:hyper-yahoo}); (ii) repeat the same procedure for $n_c,n_D,n_{re},f_{in}$ as that for $n_{es}$.

From Figure \ref{fig:hyper-ml} and  \ref{fig:hyper-yahoo}, our proposed algorithm is not  sensitive to the hyper-parameters $n_{es},n_{re},f_{in}$, except $n_c$ and $n_D$,  the number of clusters in replay buffer of the PDWA slave sampler and the iterations of learning from demonstrations each time, which also displays the importance of our prioritized experience replay mechanism with action clustering and the learning from demonstrations technique.

\subsubsection{Experiments for cascading bandits}
\begin{figure}[H]
    \centering
    \includegraphics[width=0.8\linewidth]{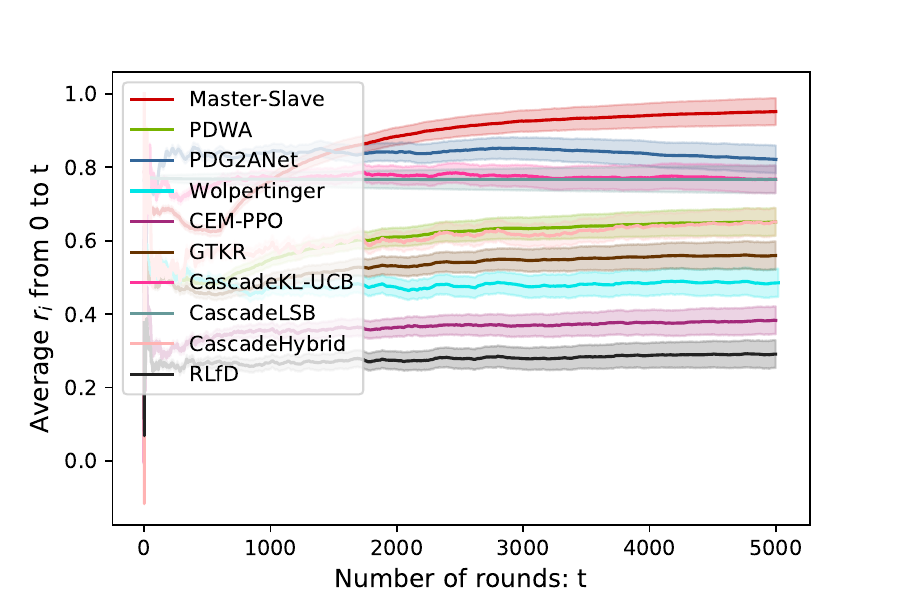}
    \caption{Cascading bandits on the  synthetic  dataset.}
    \label{fig:cascade}
\end{figure}

In this part, we test our proposed algorithm on
other online combinatorial optimization problems, e.g., the cascading bandits problem (see the discussion section in the supplementary material for the detailed description).
We follow the cascading setting in \cite{kveton2015cascading}, where  the cascade model is characterized by the attraction probabilities $\rho^{\prime}\in [0,1]^L$,  the satisfaction probabilities $v^{\prime}\in [0,1]^L$, and the discount factor $\gamma \in (0, 1]$. In the cascade model, the recommendation system recommends  a list of $K$ items $a_t = \{a_{t,1},\cdots , a_{t,K}\}$ at time $t$, and the served user examines the list from $a_{t,1}$ to  $a_{t,K}$. 
After  being examined, $a_{t,k}$  attracts the served user with probability $\rho^{\prime} (a_{t,k})$. If  $a_{t,k}$  attracts the user successfully, it will be clicked and then satisfy the user  with probability $v^{\prime}(a_{t,k})$. If $a_{t,k}$  satisfies the user successfully, the examination process will stop abruptly and the next round will begin; otherwise, item $a_{t,k+1}$ will be examined next with probability $\gamma$ and 
experience the same process as that for item $a_{t,k}$. \cite{kveton2015cascading} set the reward $r(a_t)$ to be $1$ if at least one item in the recommended list satisfies the user, and $0$ if not. 

In our implementation, we set $L=100, K=10, \gamma=0.9$ and generate $\{\rho^{\prime}(1),\cdots,\rho^{\prime}(L)\}$ and $\{v^{\prime}(1),\cdots,v^{\prime}(L)\}$ from $[0,1]^L$ uniformly at random. Then, we
perform the adaptation for our master-slave architecture stated in Section \ref{Cascading}. As shown in Figure \ref{fig:cascade} and Table \ref{tab:constraint}, our master-slave algorithm converges the fastest and enjoys the best average $r_t$ and $c_t$ all the time.

\begin{figure*}[ht] 
\centering
\includegraphics[width=0.24\linewidth]{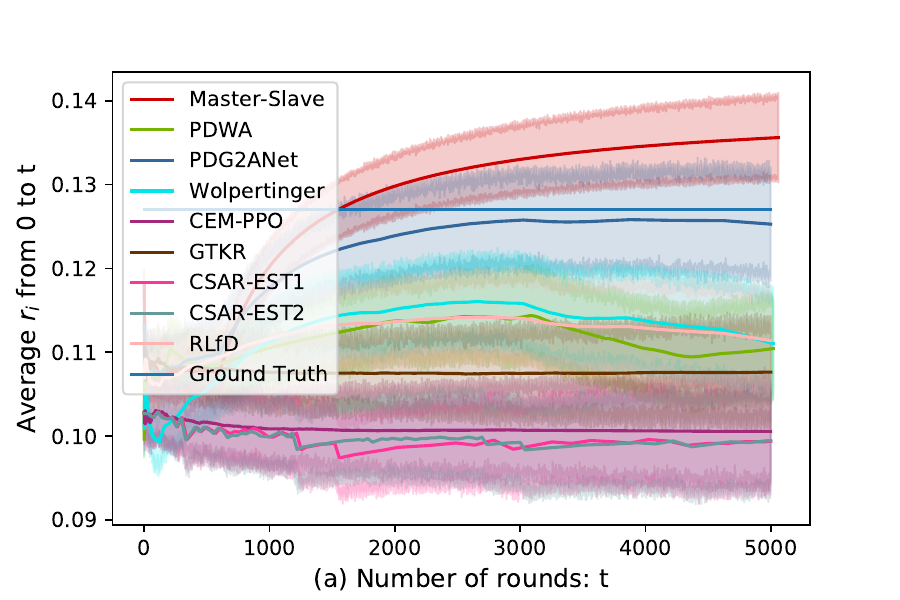}\!\!\!
\includegraphics[width=0.24\linewidth]{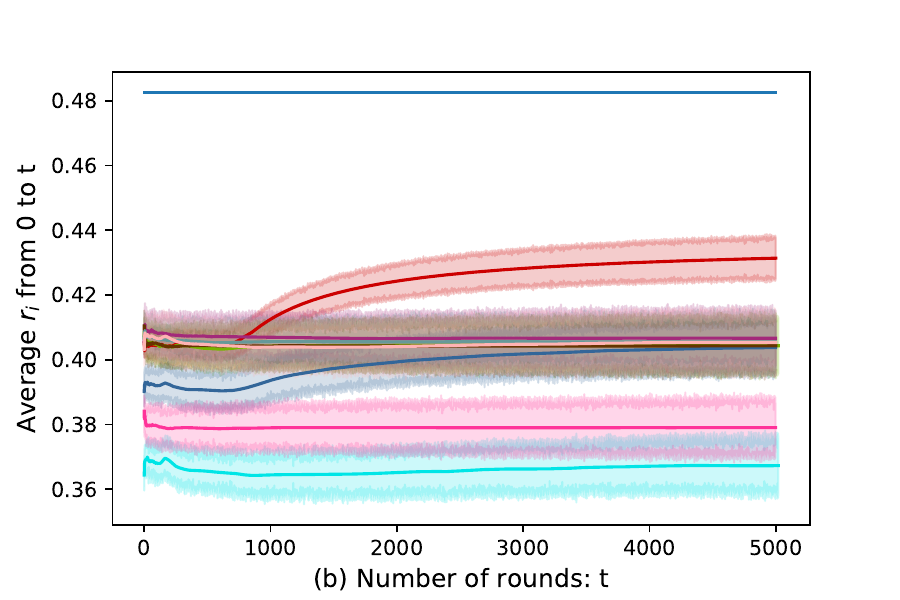}\!\!\!
\includegraphics[width=0.24\linewidth]{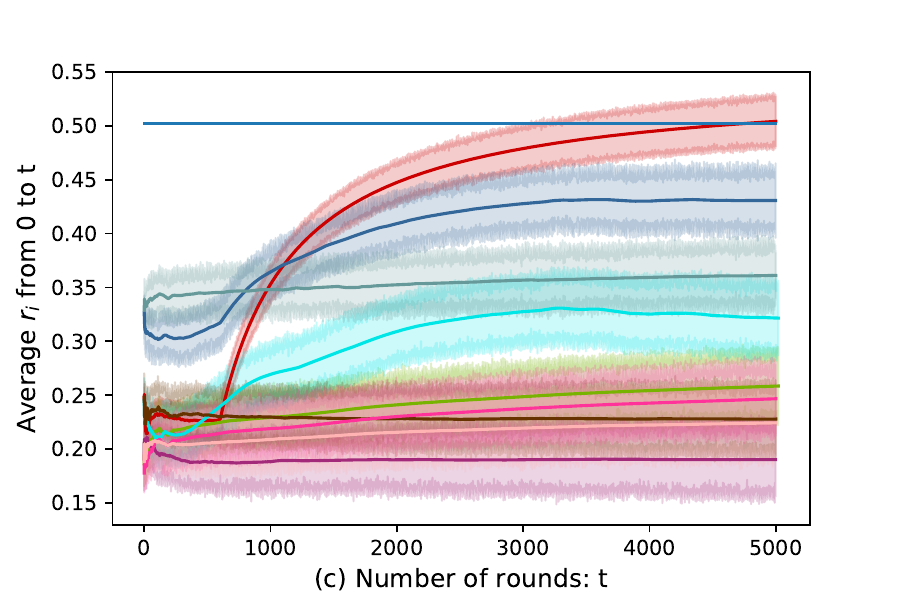}\!\!\! 
\includegraphics[width=0.24\linewidth]{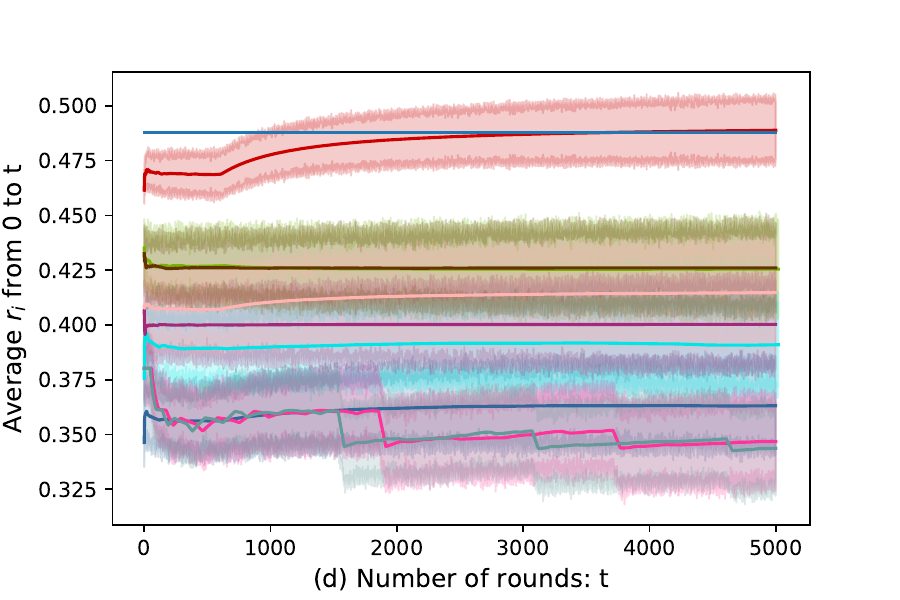} 
\caption{Synthetic datasets: (a) Linear form; (b) Quadratic form; (c) Cubic form; (d) Mixed form.}
\label{fig:Synthetic dataset}
\end{figure*}

\begin{figure*}
\centering
\includegraphics[width=0.24\linewidth]{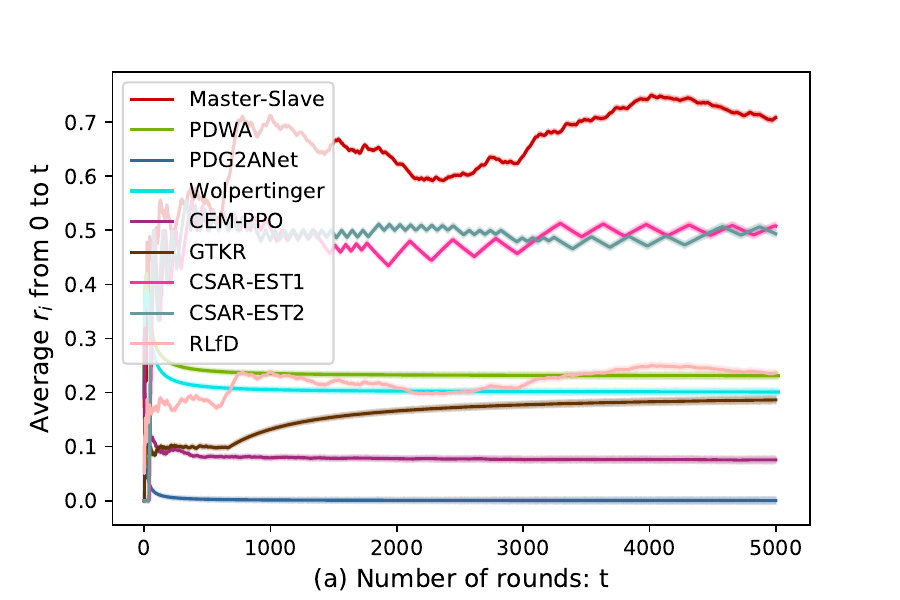}\!\!\!  
\includegraphics[width=0.24\linewidth]{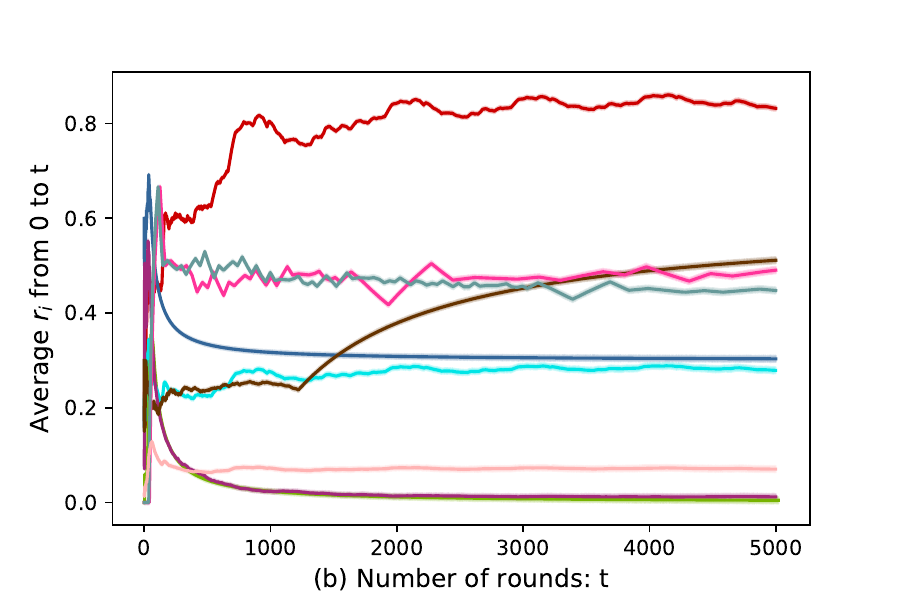} \!\!\!
\includegraphics[width=0.24\linewidth]{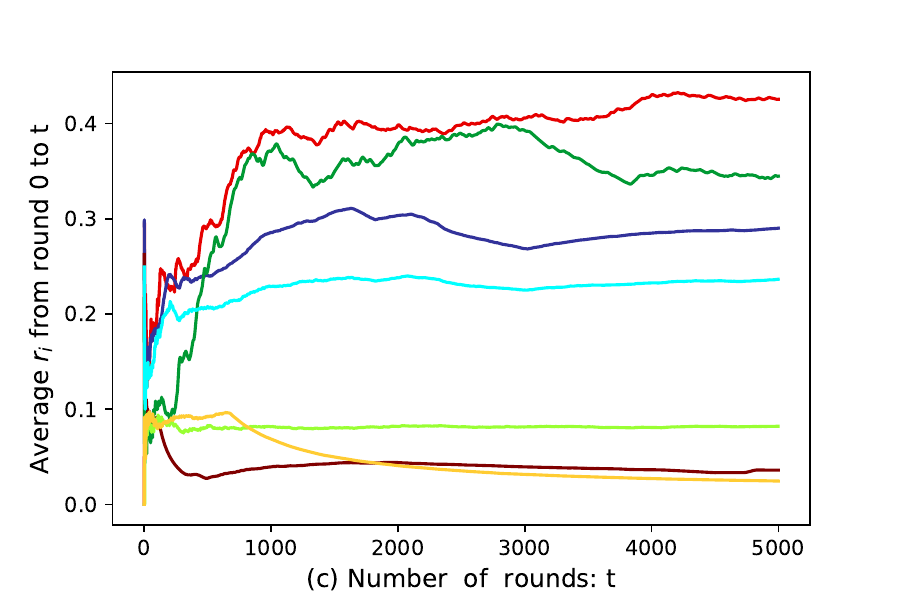}  \!\!\!
\includegraphics[width=0.24\linewidth]{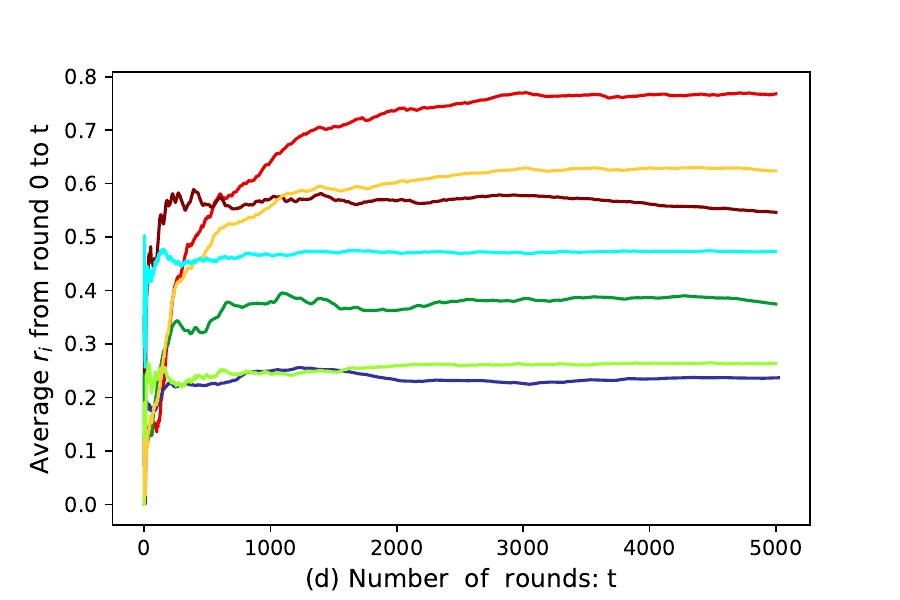}
  \caption{Real-world datasets with real feedback: (a) Delicious; (b) LastFM; (c) MovieLens; (d) Yahoo!.}
  \label{fig:Real-world datasets}
\end{figure*}

\begin{figure*}[th!] 

\centering
\includegraphics[width=0.24\linewidth]{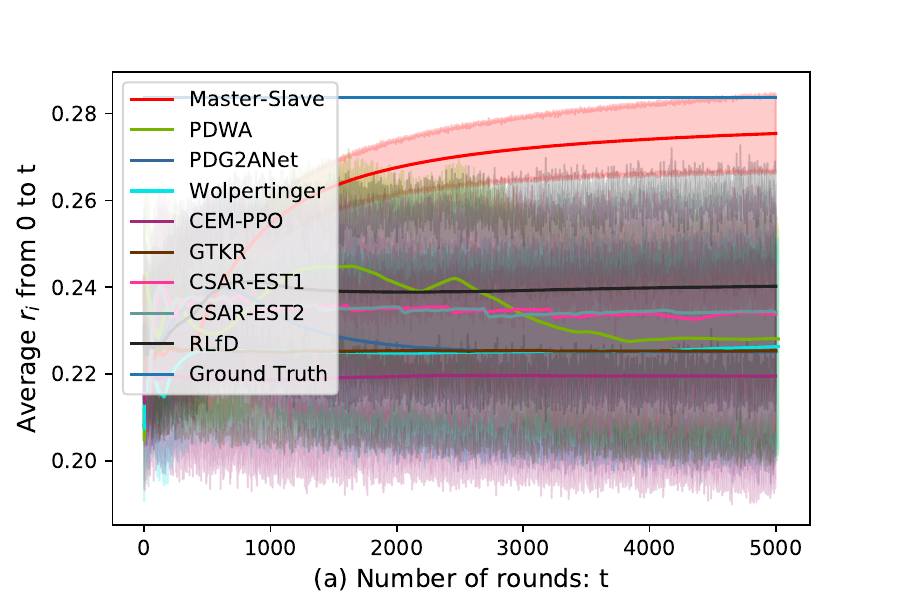} \!\!\!
\includegraphics[width=0.24\linewidth]{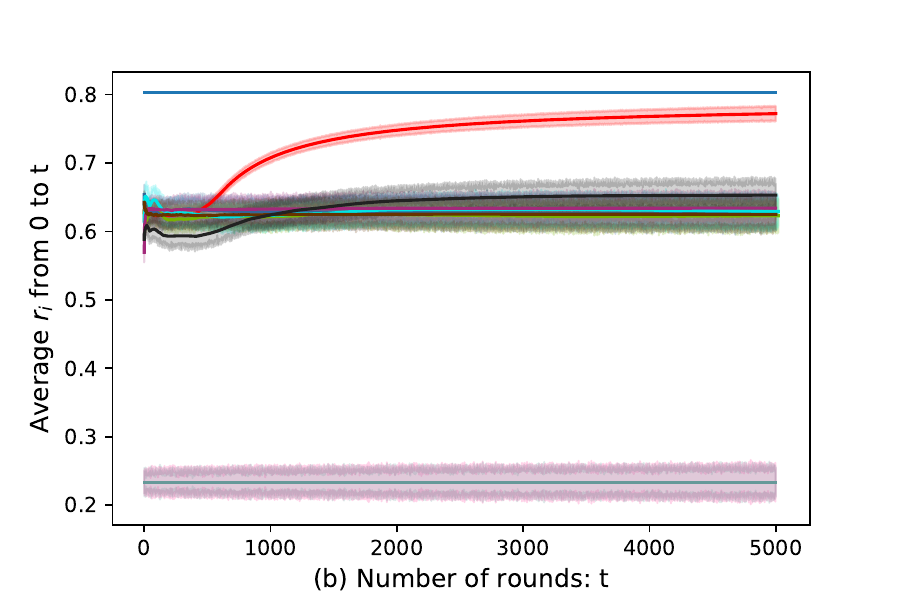} \!\!\!
\includegraphics[width=0.24\linewidth]{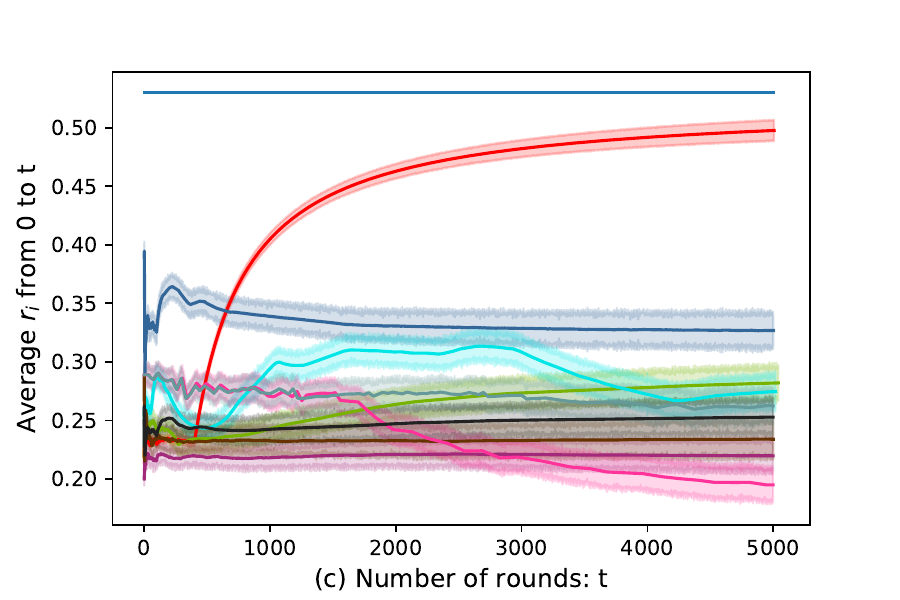} \!\!\!
\includegraphics[width=0.24\linewidth]{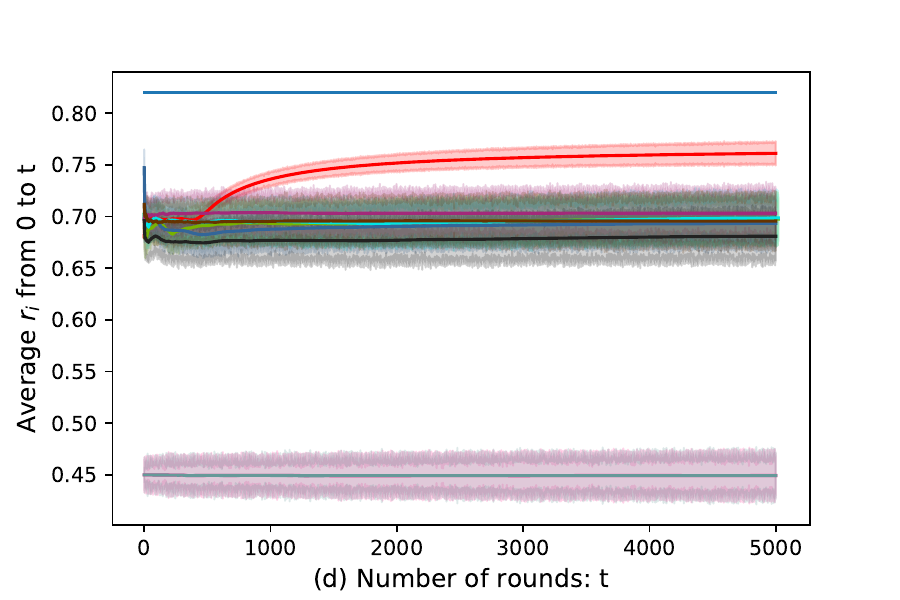} 
  \caption{MovieLens with synthetic feedback: (a) Linear form; (b) Quadratic form; (c) Cubic form; (d) Mixed form.}
 \label{fig:MovieLens dataset}
\end{figure*}

\begin{figure*}[th!] 
 \centering
\includegraphics[width=0.24\linewidth]{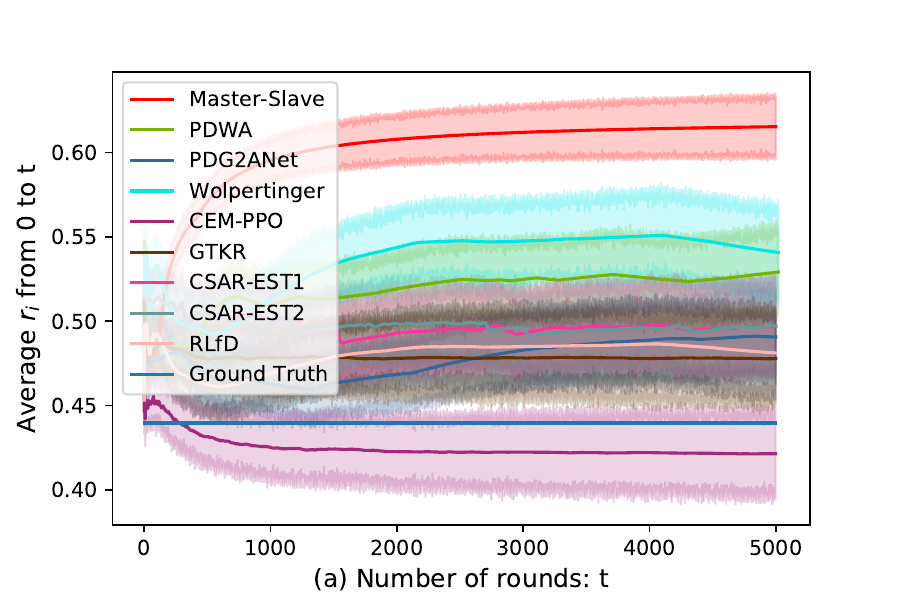} \!\!\!
\includegraphics[width=0.24\linewidth]{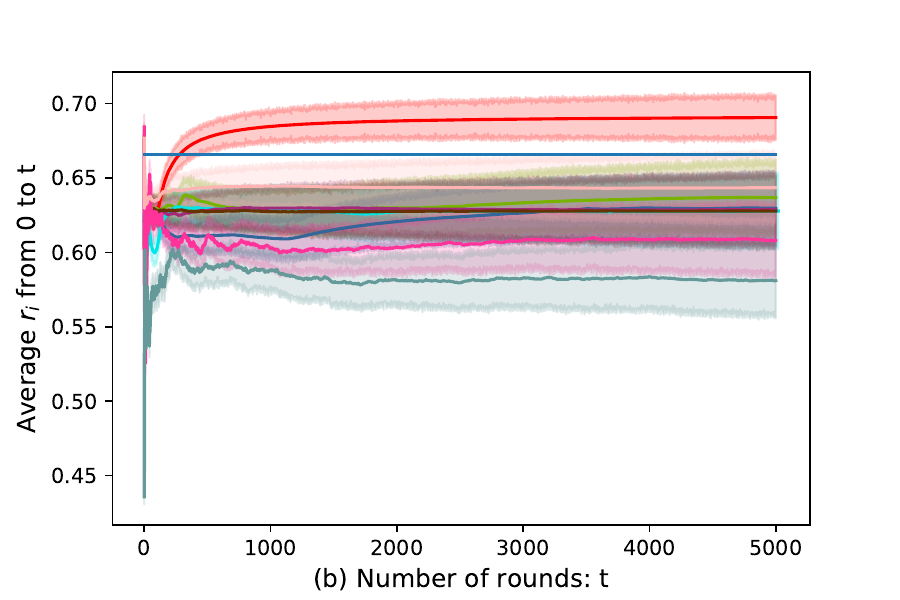}\!\!\!
\includegraphics[width=0.24\linewidth]{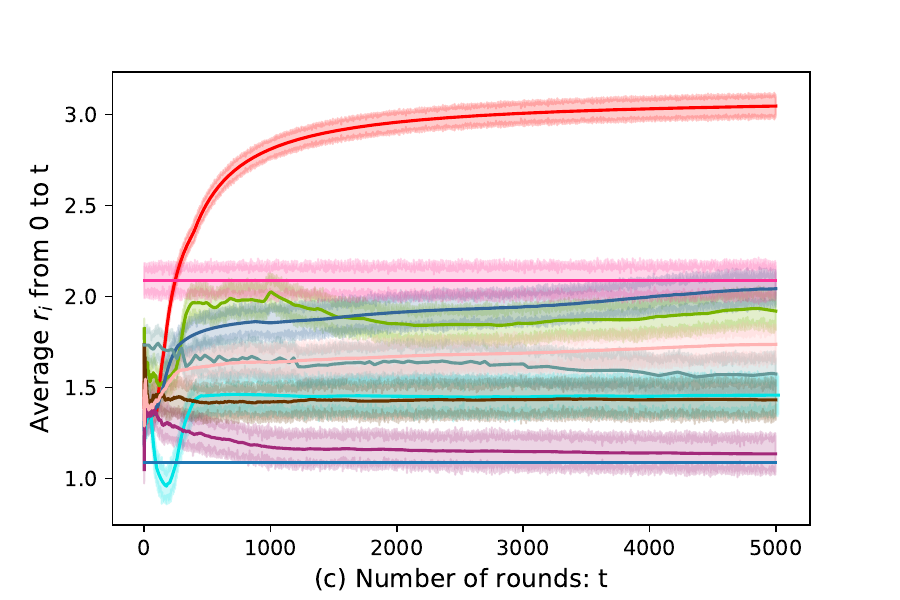} \!\!\!
\includegraphics[width=0.24\linewidth]{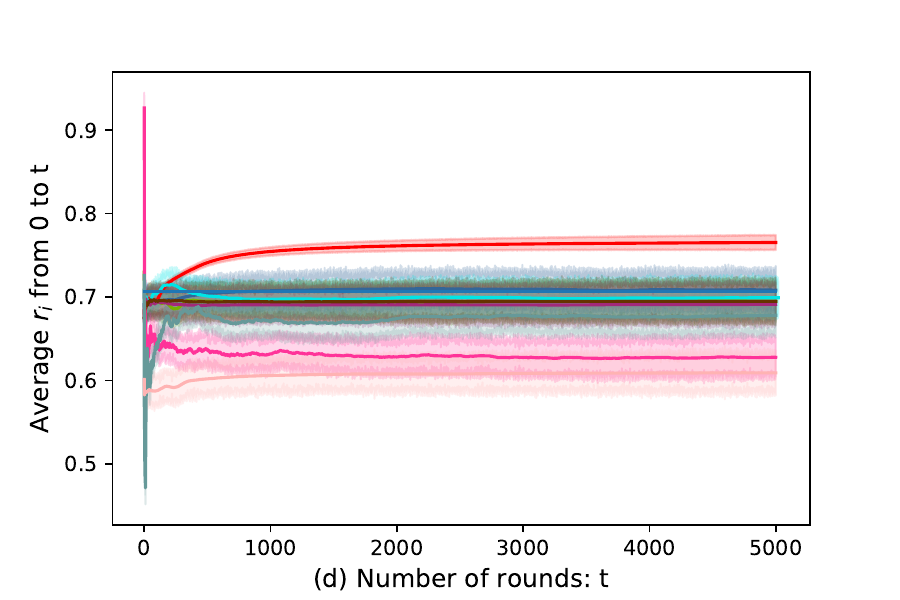}
  \caption{Yahoo! with synthetic feedback: (a) Linear form; (b) Quadratic form; (c) Cubic form; (d) Mixed form.}
 \label{fig:yahoo dataset}
\end{figure*}


 \begin{table*}[t!]
 \scriptsize
 \begin{center}
 \caption{Average constraint violation rate over $5k$ rounds.}
 \label{tab:constraint}
     \begin{tabular}{lllllllll}
 \toprule
               & \!\!\! Syn (a) \!\!\!    &\!\!\!  Syn (b) \!\!\!  & \!\!\! Syn (c) \!\!\! &\!\!\! \!\!\! Syn (d) \!\!\! & \!\!\! Delicious \!\!\! &\!\!\!  LastFM \!\!\! &\!\!\!\!\! MovieLens \!\!\!\!\! & \!\!\! Yahoo! \!\!\!  \\ \hline
\!\!\!\!\! Master-slave \!\!\! & $\mathbf{0.009}$ & $\mathbf{0.013}$ & $\mathbf{0.011}$ &$\mathbf{0.018}$ & $\mathbf{0.006}$ & $\mathbf{0.008}$& $\mathbf{0.012}$ & $\mathbf{0.010}$   \\ 
PDWA & $0.022$ & $0.039$ & $0.083$ &$0.076$ & $0.043$ & $0.071$& $0.065$ & $0.052$  \\ 
PDG2ANet & $0.023$ & $0.029$ & $0.081$ &$0.068$ & $0.038$ & $0.063$& $0.053$ & $0.057$  \\ 
\!\!\!\!\! Wolpertinger \!\!\!  & $0.039$ & $0.043$ & $0.070$ &$0.056$ & $0.023$ & $0.044$& $0.047$ & $0.041$    \\ 
CEM-PPO & $0.034$ & $0.053$ & $0.073$ &$0.071$ & $0.029$ & $0.042$& $0.039$ & $0.050$    \\ 
GTKR & $0.033$ & $0.039$ & $0.067$ &$0.048$ & $0.032$ & $0.040$& $0.041$ & $0.053$    \\
RLfD & $0.022$ & $0.047$ & $0.052$ &$0.053$ & $0.071$ & $0.046$& $0.082$ & $0.036$    \\ 
CSAR-EST1 & $0.026$ & $0.040$ & $0.049$ &$0.051$ & $0.067$ & $0.049$& $0.077$ & $0.038$   \\ 
CSAR-EST2 & $0.024$ & $0.043$ & $0.046$ &$0.054$ & $0.076$ & $0.041$& $0.069$ & $0.033$   \\   
 \bottomrule
 \end{tabular}
 \end{center}
 \end{table*}

 \begin{table*}[t!]
 \scriptsize
 \begin{center}
 \caption{$r_t$ and $c_t$ of each algorithm  in ablation study.}
 \label{tab:ablation_study_reward}
     \begin{tabular}{c|ccccccccc}
 \toprule
         Method \textbackslash{} Reward       & \!\!\! Syn (a) \!\!\!    &\!\!\!  Syn (b) \!\!\!  & \!\!\! Syn (c) \!\!\! &\!\!\! \!\!\! Syn (d) \!\!\! & \!\!\! Delicious \!\!\! &\!\!\!  LastFM \!\!\! &\!\!\!\!\! MovieLens \!\!\!\!\! & \!\!\! Yahoo! \!\!\! &\!\!\!\!\!\\ \hline
\!\!\!\!\! Master-slave \!\!\! & $\mathbf{0.1364}$ & $\mathbf{0.4308}$ & $\mathbf{0.5021}$ &$\mathbf{0.4877}$ & $\mathbf{0.7092}$ & $\mathbf{0.8125}$& $\mathbf{0.4193}$ & $\mathbf{0.7646}$  \\ 
Master-slave w/o slave 1  & $0.1308$ & $0.4189$ & $0.4795$ &$0.4638$ & $0.6493$ & $0.6922$& $0.3724$ & $0.7038$  \\ 
Master-slave w/o slave 2  & $0.1298$ & $0.4201$ & $0.4842$ &$0.4785$ & $0.6951$ & $0.7539$& $0.3891$ & $0.7202$  \\ 
Master-slave w/o slave 3  & $0.1287$ & $0.4248$ & $0.4796$ &$0.4761$ & $0.6996$ & $0.7713$& $0.3718$ & $0.7319$  \\ 
Master-slave w/o slave 4  & $0.1319$ & $0.4261$ & $0.4897$ &$0.4619$ & $0.6998$ & $0.7804$& $0.3911$ & $0.7336$  \\ 
Master-slave w/o slave 5  & $0.1249$ & $0.4177$ & $0.4641$ &$0.4492$ & $0.6344$ & $0.5957$& $0.3897$ & $0.6461$  \\ 
Master-slave w/o slave 6  & $0.1305$ & $0.4261$ & $0.4902$ &$0.4775$ & $0.6865$ & $0.7613$& $0.3881$ & $0.7268$  \\ 
 \midrule
      Method \textbackslash{} Constraint       & \!\!\! Syn (a) \!\!\!    &\!\!\!  Syn (b) \!\!\!  & \!\!\! Syn (c) \!\!\! &\!\!\! \!\!\! Syn (d) \!\!\! & \!\!\! Delicious \!\!\! &\!\!\!  LastFM \!\!\! &\!\!\!\!\! MovieLens \!\!\!\!\! & \!\!\! Yahoo! \!\!\! &\!\!\!\!\!\\ \midrule
\!\!\!\!\! Master-slave \!\!\! & $\mathbf{0.009}$ & $\mathbf{0.013}$ & $\mathbf{0.011}$ &$\mathbf{0.018}$ & $\mathbf{0.006}$ & $\mathbf{0.008}$& $\mathbf{0.012}$ & $\mathbf{0.010}$  \\ 
Master-slave w/o slave 1  & $0.018$ & $0.022$ & $0.024$ &$0.029$ & $0.019$ & $0.021$& $0.023$ & $0.022$  \\ 
Master-slave w/o slave 2  & $0.018$ & $0.022$ & $0.024$ &$0.029$ & $0.019$ & $0.021$& $0.023$ & $0.022$  \\ 
Master-slave w/o slave 3  & $0.014$ & $0.015$ & $0.017$ &$0.029$ & $0.012$ & $0.012$& $0.015$ & $0.017$  \\ 
Master-slave w/o slave 4  & $0.012$ & $0.015$ & $0.013$ &$0.020$ & $0.009$ & $0.009$& $0.014$ & $0.013$  \\ 
Master-slave w/o slave 5  & $0.018$ & $0.022$ & $0.024$ &$0.029$ & $0.019$ & $0.021$& $0.023$ & $0.022$  \\ 
Master-slave w/o slave 6  & $0.013$ & $0.015$ & $0.016$ &$0.022$ & $0.010$ & $0.011$& $0.014$ & $0.015$  \\ 
 \bottomrule
\end{tabular}
\end{center}
\end{table*}

 \begin{table}[th]
 \caption{Statistics of the LastFM, Delicious, MovieLens and Yahoo! datasets.}
 \label{tab:realdata}
 \begin{center}
     \begin{tabular}{c|cccc}
 \toprule
                    & LastFM   & Delicious & MovieLens & Yahoo!  \\ \hline
number of users & $1,892$  & $1,867$    & $2,113$   & $290$      \\ 
number of items & $17,632$ & $69,226$  & $10,197$  & $20$    \\
number of logs  & $186,480$         & $437,594$      & $47,957$   &  $128,960$ \\
 \bottomrule
 \end{tabular}
 \end{center}
 \end{table}

 \begin{table}[t!]
 \scriptsize
 \begin{center}
    \caption{$r_t$ and $c_t$ of each algorithm in the synthetic dataset with linear feedback under different $\lambda$.}
\label{tab:syn_lambda}
     \begin{tabular}{lllllll}
 \toprule
        
       Method\textbackslash{}Reward     & $\lambda$=0.1  & $\lambda$=0.5  & $\lambda$=1    & $\lambda$=5    & $\lambda$=10   & $\lambda$=50 
    \\ \midrule
Master-slave & \textbf{0.1523} & \textbf{0.1479} & \textbf{0.1466} & \textbf{0.1364} & \textbf{0.1357} & \textbf{0.1129} \\
PDWA         & 0.1204 & 0.1172 & 0.1159 & 0.1098 & 0.1071 & 0.0917 \\
PDG2ANet     & 0.1397 & 0.1324 & 0.1301 & 0.1259 & 0.1229 & 0.1094 \\
Wolpertinger & 0.1276 & 0.1218 & 0.1193 & 0.1103 & 0.1072 & 0.0995 \\
CEM-PPO      & 0.1185 & 0.1127 & 0.1103 & 0.1009 & 0.0988 & 0.0901 \\
GTKR         & 0.1237 & 0.1159 & 0.1124 & 0.1082 & 0.1068 & 0.0992 \\
RLfD         & 0.1281 & 0.1198 & 0.1172 & 0.1121 & 0.1105 & 0.1027 \\
CSAR-EST1    & 0.1143 & 0.1086 & 0.1063 & 0.0997 & 0.0969 & 0.0903 \\
CSAR-EST2    & 0.1132 & 0.1063 & 0.1042 & 0.0998 & 0.0971 & 0.0906 \\ 
 \midrule
      Method\textbackslash{}Constraint       & $\lambda$=0.1  & $\lambda$=0.5  & $\lambda$=1    & $\lambda$=5    & $\lambda$=10   & $\lambda$=50 
    \\ \midrule
    Master-slave & \textbf{0.021} & \textbf{0.015} & \textbf{0.013} & \textbf{0.009} & \textbf{0.008} & \textbf{0.004} \\
PDWA         & 0.036 & 0.029 & 0.027 & 0.022 & 0.021 & 0.017 \\
PDG2ANet     & 0.039 & 0.033 & 0.029 & 0.023 & 0.021 & 0.018 \\
Wolpertinger & 0.056 & 0.049 & 0.048 & 0.039 & 0.037 & 0.031 \\
CEM-PPO      & 0.049 & 0.044 & 0.042 & 0.034 & 0.031 & 0.025 \\
GTKR         & 0.048 & 0.042 & 0.039 & 0.033 & 0.031 & 0.023 \\
RLfD         & 0.037 & 0.029 & 0.028 & 0.022 & 0.020 & 0.017 \\
CSAR-EST1    & 0.041 & 0.034 & 0.032 & 0.026 & 0.025 & 0.021 \\
CSAR-EST2    & 0.040 & 0.032 & 0.031 & 0.024 & 0.022 & 0.019 \\ 
 \bottomrule
 \end{tabular}
 \end{center}
\end{table}

 \begin{table}[t!]
 \scriptsize
 \begin{center}
    \caption{$r_t$ and $c_t$ of each algorithm in the delicious dataset  under different $\lambda$.}
\label{tab:del_lambda}
     \begin{tabular}{lllllll}
 \toprule
        
       Method \textbackslash{} Reward     & $\lambda$=0.1  & $\lambda$=0.5  & $\lambda$=1    & $\lambda$=5    & $\lambda$=10   & $\lambda$=50 
    \\ \hline
Master-slave & \textbf{0.7876} & \textbf{0.7422} & \textbf{0.7383} & \textbf{0.7092} & \textbf{0.7052} & \textbf{0.6447} \\
PDWA         & 0.3024 & 0.2871 & 0.2746 & 0.2496 & 0.2366 & 0.2005 \\
PDG2ANet     & 0.2924 & 0.2681 & 0.2637 & 0.0032 & 0.1972 & 0.1336 \\
Wolpertinger & 0.2784 & 0.2495 & 0.2402 & 0.2093 & 0.1938 & 0.1699 \\
CEM-PPO      & 0.1877 & 0.1591 & 0.1489 & 0.0891 & 0.1182 & 0.0821 \\
GTKR         & 0.2869 & 0.2428 & 0.2353 & 0.1912 & 0.1831 & 0.1575 \\
RLfD         & 0.3387 & 0.3095 & 0.2902 & 0.2497 & 0.2386 & 0.2097 \\
CSAR-EST1    & 0.5896 & 0.5677 & 0.5601 & 0.5032 & 0.4892 & 0.4461 \\
CSAR-EST2    & 0.5803 & 0.5415 & 0.5302 & 0.4997 & 0.4783 & 0.4296 \\ 
 \hline
      Method \textbackslash{} Constraint       & $\lambda$=0.1  & $\lambda$=0.5  & $\lambda$=1    & $\lambda$=5    & $\lambda$=10   & $\lambda$=50 
    \\ \hline
Master-slave & \textbf{0.017} & \textbf{0.011} & \textbf{0.009} & \textbf{0.006} & \textbf{0.005} & \textbf{0.004} \\
PDWA         & 0.054 & 0.049 & 0.048 & 0.043 & 0.04  & 0.031 \\
PDG2ANet     & 0.052 & 0.046 & 0.044 & 0.038 & 0.036 & 0.028 \\
Wolpertinger & 0.035 & 0.029 & 0.028 & 0.023 & 0.022 & 0.015 \\
CEM-PPO      & 0.043 & 0.036 & 0.034 & 0.029 & 0.028 & 0.022 \\
GTKR         & 0.045 & 0.040 & 0.039 & 0.032 & 0.029 & 0.021 \\
RLfD         & 0.078 & 0.075 & 0.074 & 0.071 & 0.066 & 0.059 \\
CSAR-EST1    & 0.076 & 0.073 & 0.072 & 0.067 & 0.063 & 0.048 \\
CSAR-EST2    & 0.082 & 0.080 & 0.079 & 0.076 & 0.071 & 0.054 \\ 
\bottomrule
\end{tabular}
\end{center}
\end{table}

\begin{table*}
\centering
\caption{Average reward of adopting each single slave sampler in both synthetic and real datasets.}
\label{tab:single}
\begin{tabular}{lllllllll}
\toprule
& Syn(a) & Syn(b) & Syn(c) & Syn(d) & Delicious & LastFM & MovieLens & Yahoo! \\\midrule
Solver Sampler                                  & 0.1207 & 0.4235 & 0.4772 & 0.4593 & 0.6392    & 0.7029 & 0.3709    & 0.6381 \\
Random Sampler with keeping the Best in History & 0.0879 & 0.3821 & 0.4721 & 0.4396 & 0.4491    & 0.5193 & 0.3193    & 0.4891 \\
G2anet Sampler                                  & 0.1192 & 0.4079 & 0.4508 & 0.4325 & 0.5529    & 0.6298 & 0.3285    & 0.5536 \\
CEM-PPO Sampler                                 & 0.1264 & 0.4184 & 0.4631 & 0.4428 & 0.6031    & 0.6731 & 0.3518    & 0.5837 \\
Primal-dual Wolpertinger Sampler                & 0.1187 & 0.4098 & 0.4599 & 0.4339 & 0.5893    & 0.6639 & 0.3329    & 0.5499 \\
\bottomrule
\end{tabular}
\end{table*}

\subsection{Details of master model}\label{master}

\textbf{Neural contextual UCB-based network.}
Neural contextual UCB-based network (NeuralUCB) was firstly introduced in \cite{zhou2019neural} to deal with non-linear reward in contextual bandits. Since we regard the action as a vector instead of a set, NeuralUCB can also adapt to our top-$K$ setting.  It can be regarded as a neural non-linearization of the Linear UCB algorithm in \cite{chu2011contextual} and approximates $h$  by a fully connected neural network with depth $L_1\geq 2$:
$$h'(x;\theta)=\sqrt{m} \mathbf{W}_{L_1} \sigma(\mathbf{W}_{L_1-1} \sigma(\cdots \sigma(\mathbf{W}_{1} x))),$$

where $\sigma (x) = \max\{x, 0\}$ is the rectified linear unit (ReLU) activation function, $\mathbf{W}_{1} \in \mathbb{R}^{m \times L}, \mathbf{W}_{i} \in \mathbb{R}^{m \times m},$ $ 2 \leq i \leq L_1-1, \mathbf{W}_{L_1} \in \mathbb{R}^{m \times 1}, \text { and } \theta=$ $[\operatorname{vec}(\mathbf{W}_{1})^{\top}, \ldots, \operatorname{vec}(\mathbf{W}_{L_1})^{\top}]^{\top} \in \mathbb{R}^{p}$ with $p=m+m L+m^{2}(L_1-1) $. At each round $t$, $\theta$ is trained on $(A_1,r_1),(A_2,r_2),\cdots,(A_{t-1},r_{t-1})$.

Denote the gradient of the neural
network function by 
$\mathbf{g}(x; \theta)=\nabla_{\theta} h'(x; \theta) \in \mathbb{R}^{p}$. Then NeuralUCB  goes as follows:

\begin{itemize}
    \item \textbf{Initialization:}
    
    \begin{itemize}
        \item Special initialization on $\theta_{0}$:
        
For $1 \leq l \leq L_1-1$, $i\in [m/2]$ and $j\in [m/2]$ if $l>1$, otherwise $j\in [L/2]$,
$$
\mathbf{W}_{l}=(\begin{array}{cc}
\mathbf{W} & \mathbf{0} \\
\mathbf{0} & \mathbf{W}
\end{array}), \mathbf{W}_{\{i, j\}} \sim N(0,4 / m).
$$
 
 For $L_1$, $i\in [m/2]$, $\mathbf{W}_{L_1}=(\mathbf{w}^{\top},-\mathbf{w}^{\top}), \mathbf{w}_{\{i\}} \sim N(0,2 / m)$.
 \item Normalization on each $x$ in the action candidate pool with its 2-norm being one.
    \end{itemize}
    \item \textbf{At each round $t$:}
    \begin{itemize}
        \item Obtain the new action candidate pool.
        \item Calculate the upper confidence bound for an action candidate $x$: 
        $$Var_{t,x}= \underbrace{\sqrt{\mathbf{g}(x; \theta_{t-1})^{\top} \mathbf{Z}_{t-1}^{-1} \mathbf{g}(x; \theta_{t-1}) / m}}_{\text {variance }}.$$
        \begin{equation}
            U_{t, x}=\underbrace{h'(x; \theta_{t-1})}_{\text {mean }}+\gamma_{t-1}Var_{t,x}.\label{eq:U}
        \end{equation}
        
        \item  Select the action $x_t$ with the highest upper confidence bound for decision.
        \item Update $\mathbf{Z}_t$ and $\theta_t$ by
        $$\mathbf{Z}_{t}=\mathbf{Z}_{t-1}+\mathbf{g}(x_{t} ; \theta_{t-1}) \mathbf{g}(x_{t} ; \theta_{t-1})^{\top} / m.$$
        Denote the loss function $\mathcal{L}(\theta)$ as
$$
\mathcal{L}(\theta)=\sum_{i=1}^{t}(h'(x_{i} ; \theta)-r_{i})^{2} / 2+m \lambda_1\|\theta-\theta^{(0)}\|_{2}^{2} / 2.
$$
Run $J$ steps of gradient descent on $\mathcal{L}(\theta)$ starting from $\theta_{0},$ and take $\theta_{t}$ as the last iterate:
$$
\theta^{(0)}=\theta_{0}, \theta^{(j+1)}=\theta^{(j)}-\eta \nabla \mathcal{L}(\theta^{(j)}), \theta_{t}=\theta^{(J)}.
$$
\item Update $\gamma_{t}$ by
\begin{align*}
    \gamma_{t}=&O(\underbrace{\sqrt{\lambda_1} S+\nu \sqrt{\log \frac{\operatorname{det} \mathbf{Z}_{t}}{\delta \operatorname{det} \lambda_1 \mathbf{I}}}}_{\text {confidence radius }}\\&+\underbrace{(\lambda_1+t L_1)(1-\eta m \lambda_1)^{J / 2} \sqrt{t / \lambda_1}}_{\text {function approximation error }}),
\end{align*}
where $S$ is the norm parameter, $\nu$ is the exploration parameter, and $\eta$ is the step size in the gradient descent procedure.
    \end{itemize}
\end{itemize}
We introduce our adaptation for NeuralUCB to handle non-stationary environments in Discussion. 
\subsection{Details of slave models}\label{slave} 
 
 \subsubsection{G2ANet sampler}
   \textbf{Game Abstraction based on Two-stage Attention Network (G2ANet).} \
  The slave models mentioned above ignore the mutual influence of  each two arms in the   $c_t$ terms. Therefore, we employ a graph attention network  Game Abstraction based on Two-stage Attention Network (G2ANet) in the multi-agent environment \cite{liu2020multi} to infer the diversity constraints between every two arms and make a comprehensive decision for each arm.
  
   In the multi-agent environment,  complex cooperation and competition  among a large number of agents make strategy learning very difficult. In addition, in the decision-making process, each agent does not need to interact with all agents all the time, but only needs to interact with its neighboring agents. Traditional methods can only determine which agents have interactions through prior knowledge. When the system is very complex, it is very difficult to define interactions based on rules. Yong Liu et al. \cite{liu2020multi} aimed to  determine whether there is an interaction between
  every two agents and if the interaction exists, to 
  estimate the importance of the interaction's influence on each agent's strategy.
   
In \cite{liu2020multi}, the multi-agent system is modeled as a graph network, which is a fully connected topology graph. Each node in the graph represents an agent, and the edge between two nodes represents the interaction between two agents. A Two-Stage Attention Neural Network is used to reason about the interaction mechanism between  agents:

\begin{itemize}
    \item Hard-attention is  designed to disconnect irrelevant interactive edges. The usual hard-attention mechanism is obtained through sampling and is not differentiable. G2ANet improves it by using a gumbel-softmax function \cite{jang2016categorical} to enable end-to-end learning.
    \item Soft-attention aims to judge the importance weights of the interactive edges retained by hard-attention.
\end{itemize}

G2ANet integrates the above two attention mechanisms with reinforcement learning (RL) algorithms such as reinforce   during the process of multi-agent strategy learning:

Consider a locally observable environment. For the agent $i$, Yong Liu et al. \cite{liu2020multi} encoded the local observation   into a  vector $h_i$ by a Multi-Layer Perception.

First, use Bi-LSTM  to implement the hard-attention mechanism so as to discover the existence of an interactive relationship between each pair of agents. For each agent pair $(i,j)$, combine the features of agent $i$ and agent $j$ to get $(h_{i}, h_{j})$. Take $(h_{i}, h_{j})$ as the input of the Bi-LSTM model and get the output $h_{i, j}=f_1(B i-\operatorname{LSTM}(h_{i}, h_{j}))$, where $f_1$ is a fully connected layer.

 Since the hard-attention mechanism involves the sampling process and cannot back-propagate gradients, G2ANet tries to use the gumbel-softmax function to deal with the non-differentiable issue, and obtains the $0-1$ hard-attention for the edge between agent $i$ and agent $j$: $$W_{h}^{i, j}=\operatorname{gum}(f_1(L \operatorname{STM}(h_{i}, h_{j}))).$$

Through the hard-attention mechanism, we can get the subgraph $G_i$ for   agent $i$, and use the soft-attention mechanism to learn the weight of each edge in the subgraph $G_i$. The weight of the edge  between agent $i$ and agent $j$ is $W_{s}^{i, j} \propto \exp (e_{j}^{T} W_{k}^{T} W_{q} e_{i} W_{h}^{i, j})$, where $e_i$ and $e_j$ are the embeddings of agent $i$ and agent $j$, respectively, $W_k$ and $W_q$ are a key linear mapping and a query linear mapping, respectively. $e_j$  is converted into a key by $W_k$, and $e_i$ is converted into a query by $W_q$.

Through the Two-Stage Attention Neural Network, we can get a reduction graph in which each agent is only connected to the agent that needs to interact. Use the weights output by the soft-attention mechanism to aggregate the neighboring features and we obtain a vector $x_i$ which represents the neighboring information of agent $i$. Finally, G2ANet applies the policy gradient algorithm REINFORECE to get the action for each agent:
$$a_{i}=\pi(h_{i}, x_{i}),$$

where $\pi$ is the action policy, $h_i$ is the observation feature for agent $i$, and $x_i$ is the contribution from other agents for agent $i$.

\textbf{Improved gumbel top-$K$ sampling.} \ 

At each round, G2ANet outputs a real number $out_i\in [0,1]$ for each node $i\in [L]$.  Based on
$\{out_i\}_{i\in [L]}$, we  adapt the    differentiable gumbel top-$K$ sampling algorithm (GTKS)
\cite{kool2019stochastic} to obtain the binary elite sample  $\{d_1,\cdots,d_L\}$ and estimate the sampling probability.

The original GTKS first 
performs the same data perturbation process as the gumbel-max sampling \cite{kool2019stochastic} on $out_i$ to obtain $out_i^{\prime}$ for $i\in [L]$. Then, draw an ordered sample of size $K$ as the top $K$ arms by $ I_{1}^{*}, \ldots, I_{K}^{*}=\arg \operatorname{top}_{K} \{out^{\prime}_{1},\cdots,out^{\prime}_{i},\cdots, out^{\prime}_{L}\} $ with $ out^{\prime}_{I_{1}^{*}} \geq out^{\prime}_{I_{2}^{*}} \geq \ldots,out^{\prime}_{I_{K}^{*}}$.  And
 calculate the ordered sampling probability $P(I_{1}^{*}\!=\!i_{1}^{*}, \ldots, I_{K}^{*}\!=\!i_{K}^{*})\!$.

However, here comes the problem that the choice of arms should be considered regardless of order during the top-$K$ sampling process, that is, the order of $out^{\prime}_{I_{1}^{*}} \geq out^{\prime}_{I_{2}^{*}} \geq \ldots,out^{\prime}_{I_{K}^{*}}$   should not be required.
  To  eliminate the above requirement, one should  perform $K!$ permutation for $I_{1}^{*}, \ldots, I_{K}^{*}$ and average the $K!$ ordered probabilities  to obtain the unordered sampling probability.
 To reduce the amount of calculation,  we only randomly generate $M$ permutations of  $I_{1}^{*}, I_{2}^{*}, \cdots, I_{K}^{*}$, and average the corresponding $M$ probabilities as the unordered sampling probability.

\textbf{Feedback after outputting a sample.} \ 
The composite feedback at each round   is calculated by $Surrogate(\cdot) -\lambda c(\cdot)$.

\subsection{Discussion}
 
 \subsubsection{Novelty}\label{novelty}
 
 Solving the  online constrained combinatorial optimization problems (O-CCOPs) can face many challenges, such as huge action space, great difficulty to satisfy constraints and handle the reward-constraint   trade-off, etc. Since we do not know much environmental information, we cannot use traditional optimization solvers to solve the above problems and need to resort to deep neural networks. However, normal 
deep methods are poor at constraint-handling and are time-consuming for training, so we modify and co-train solvers and deep (reinforcement) learning methods to face the above challenges.  Specifically, we
design a novel master-slave hierarchical architecture
which leverages the cooperation of learning, sampling, and traditional optimization methods,
to efficiently  explore the combinatorial and constrained action space.

 For the master model, we  utilize a neural contextual UCB-based network  to estimate the user feedback, provide surrogate rewards to train slave samplers, and estimate  samples collected from slave models to make a decision. 
 Apart from the master model, we deliberately design multiple slave models, each of which has its  distinguishing merit to deal with the  proposed setting:

\textbf{(i)} 
Solver sampler finds the solutions that strictly satisfy all constraints but may not own high rewards.
    \textbf{(ii)} Primal-dual Wolpertinger sampler can efficiently search in the combinatorial space by designing proto-action and balance the reward-constraint tradeoff by primal-dual-based reward shaping.
    \textbf{(iii)} G2anet sampler makes up for the pity that the above samplers do not thoroughly study the influence of constraints on each pair of variables.
    \textbf{(iv)} Both Wolpertinger and G2anet require time-consuming backpropagation. In contrast, CEM-PPO sampler requires much less of it, can be naturally used for various O-CCOPs, is easier to scale in distributed environments, and has fewer hyper-parameters.
    \textbf{(v)} Teacher-student sampler promotes samplers to learn from each other and generate more diversified and elite samples. 
During the sample generating process, we also propose  policy co-training to avoid local optimal solutions and achieve a mutual benefit among  slave models. 

As to other highlights of our method,
    \textbf{(i)}  we  design a prioritized replay buffer with action clustering and reward-constraint hierarchies to boost the diversity of elite samples, the accurate approximation of the critic network, and the in-time adjustment in face of the reward-constraint imbalance;
    \textbf{(ii)} we apply G2anet to explore the constraint relationship between arms from a multi-agent interaction perspective;
    \textbf{(iii)} we adapt the differentiable gumbel top-K sampling trick to fit in our setting and improve G2anet and CEM-PPO;
    \textbf{(iv)} a policy co-training mechanism is applied to avoid local optima and facilitate different samplers to learn from each other;

 In brief, we establish a general framework for O-CCOPs and discuss its modifications for non-stationary scenarios and other O-CCOPs. Experiments on both synthetic and real-world datasets for constrained top-$K$ bandits and     cascading bandits  demonstrate the superiority of our  framework.

\subsubsection{Some clarifications}\label{clarify}
 \textbf{(i) Calculation of normalized edit distance (NED).}  In our experiments, we calculate NED by the following steps: Given two equal-length numerical vectors $[a_1,a_2,\dots,a_n]$ and $[b_1, b_2,\dots,b_n]$.
Compute the absolute difference $d_i= |a_i - b_i|, \forall i \in [1,2,\cdots,n]$ between the elements corresponding to the two vectors.
Then the Normalized Edit Distance between $[a_1,a_2,\dots,a_n]$  and $[b_1, b_2,\dots,b_n]$ is calculated by $NED = \frac{\sum_{i=1}^{n}{d_i}}{\sum_{i=1}^{n}{a_i} + \sum_{i=1}^{n}{b_i}}$.

 \textbf{(ii) Concepts of elite samples.}
 Samples refer to $L$-sized vectors generated by the slave samplers.

 Elite samples refer to samples that are generated  by  slave samplers and provided to the master.
 
 \textbf{(iii) Linear pair between reward and constraint.} 
For simplicity, 
in our proposed algorithm we customize the surrogate feedback for the slave models by linearly combining the reward and constraint violation rate. In spite of this linear design, our algorithm enjoys 
both the best reward and the best constraint violation rate in experiments due to its elaborate design of the master and slave models,  and therefore could also perform the best with regard to any composite metric  monotonous to reward and constraint violation rate.

 \textbf{(iv) The difficulty of constraint handling.}
Note that in combinatorial optimization problems,
the difficulty and importance of constraint-handling are nonnegligible.  Actually, in many cases of integer programming, the difficulty of solving (un-)constrained problems can be worlds apart. And the imperfect nature of real recommendation makes constrained decisions ubiquitous.

 \textbf{(v) Settings with and without context.}
 Our algorithm can adapt to both contextual bandits and non-contextual bandits by deciding whether to use the g2anet sampler or not and 
re-designing the state of the RL-based samplers.

\textbf{(vi) Sample assignment.} In the experimental evaluation, we calculate the percentage of elite samples that each slave model should provide  by  performing the softmax function on the average surrogate score of 
historical samples each slave model outputs. Specifically,
 denote the average surrogate score of historical samples the slave model $i$ outputs as $s_{i}$, and then the percentage of elite samples that the slave model $i$ should provide is set to be $\frac{\exp{s_i}}{\sum_{j=1}^{6}\exp{s_j}}.$ 
 
  \textbf{(vii) Ensemble learners.}
 Although many good ensemble learners exist today, they mainly work for prediction rather than for decision. Therefore, bagging-based and boosting-based  ensemble learners are not comparable to our master-slave architecture. 

 As to the existing master-slave architecture in RL such as \cite{wu2018master,kong2017revisiting},
 our framework is similar to that in those papers, but the functions of the master and slaves, the interaction mode, and the cooperation mechanism among slaves are different. Moreover, many additional details are considered to  improve the overall quality and diversity of all global elite samples as much as possible and search the constrained combinatorial action space as efficiently as possible.
 
\textbf{(viii) Time complexity.}
 We test the elapsed time of each sampler by carrying out our master-slave architecture with each single slave. As can be seen in Table \ref{tab:time},
except the random sampler, other samplers do not differ much with regard to the magnitude of elapsed time. Among all samplers,
the G2anet sampler consumes the most time and the CEM-PPO sampler takes up the second place. The solver sampler ranks medium. The Primal-dual Wolpertinger sampler and the random sampler consume the least time. 
In fact,
the interaction frequency between the master model and each slave sampler is adjustable according to the
computation budget of each slave sampler. For example, some samplers might require time-consuming back-propagation and are
sample inefficient. For these samplers, we could decrease their interaction frequency with the master
properly. Also, some slave samplers can be discarded if the time and computation resources are extremely
constrained. 

For the comparison of the computational complexity of our algorithm with other baselines (GTKR, RLfD, CSAR-EST1, CSAR-EST2\footnote{We update the slave samplers of our master-slave algorithm, the network of GTKR and RLfD according to the replay buffer every $f_{in}$ times. Meanwhile, we use multiprocessing with $6$ processes to run our $6$ slave samplers at the same time during each update.}), as can be seen in Table \ref{tab:time}, although our master-slave architecture owns the longest running time, its wall-clock time is not dramatically longer than other deep reinforcement learning baselines. In the future, we could speed up the evolution of each slave sampler by increasing the frequency of the policy co-training technique, so that our proposed algorithm could produce higher quality solutions in less time.

  \textbf{(ix) Theoretical guarantee.}  Zhou et al.   proved a near-optimal regret guarantee for NeuralUCB, the core component of our master algorithm. See \cite{zhou2019neural} for the detailed derivation of NeuralUCB.

\begin{table*}
\centering
\tiny
\caption{Elapsed time of adopting each sampler, all samplers, and other baselines.}\label{tab:time}
\begin{tabular}{lllll}\toprule
                                           Time (minute)     & Mean time in synthetic datasets & Std time in synthetic datasets & Mean time in real-world datasets & Std time in real-world datasets \\\midrule
Solver Sampler                                  & 25.61                           & 2.97                           & 10.42                            & 1.53                            \\
Random Sampler with keeping the Best in History & 4.92                            & 0.61                           & 2.05                             & 0.26                            \\
G2anet Sampler                                  & 51.04                           & 3.89                           & 21.46                            & 2.86                            \\
CEM-PPO Sampler                                 & 27.42                           & 2.72                           & 12.38                            & 1.64                            \\
Primal-dual Wolpertinger Sampler                & 20.04                           & 1.96                           & 9.34                             & 1.72                            \\
Master-slave                                    & 70.28                           & 5.06                           & 37.95                            & 3.85                           \\
GTKR& 21.47& 2.01& 11.83& 1.68                        \\
RLfD&52.41& 3.81&23.08& 2.76  \\
CSAR-EST1&10.28&0.83&5.03&0.49   \\
CSAR-EST2&10.46&0.85&5.29&0.51  \\
\bottomrule
\end{tabular}
\end{table*}
  
  \subsubsection{Non-stationary user feedback}\label{nonstationary}
 When the feedback function $h$ is time-varying, we can combine NeuralUCB  with the discounted linear upper confidence bound algorithm \cite{russac2019weighted} to deal with the non-stationary issue, that is,
\begin{itemize}
  \item \textbf{Initialization:}
    
    \begin{itemize}
        \item  Special initialization on $\theta_{0}$:
        
 For $1 \leq l \leq L_1-1$, $i\in [m/2]$ and $j\in [m/2]$ if $l>1$ otherwise $j\in [L/2]$,
$$
\mathbf{W}_{l}=(\begin{array}{cc}
\mathbf{W} & \mathbf{0} \\
\mathbf{0} & \mathbf{W}
\end{array}), \mathbf{W}_{\{i, j\}} \sim N(0,4 / m).
$$
 
 For $L_1, i\in [m/2]$, $\mathbf{W}_{L_1}=(\mathbf{w}^{\top},-\mathbf{w}^{\top}), \mathbf{w}_{\{i\}} \sim N(0,2 / m)$.
 \item  Normalization on each $x$ in the action candidate pool with its 2-norm being one.
    \end{itemize}
    \item  \textbf{At each round $t$:}
    \begin{itemize}
        \item  Obtain the new action candidate pool.
      \item  Calculate the upper confidence bound for an action candidate $x$: 
        $$Var_{t,x}= \underbrace{\sqrt{\mathbf{g}(x; \theta_{t-1})^{\top} \mathbf{Z}_{t-1}^{-1}\tilde{\mathbf{Z}}_{t-1} \mathbf{Z}_{t-1}^{-1} \mathbf{g}(x; \theta_{t-1}) / m}}_{\text {variance }}.$$
        $$U_{t, x}=\underbrace{h'(x; \theta_{t-1})}_{\text {mean }}+\alpha_{t-1}Var_{t,x},$$
        where $\alpha_{t-1}$ is a time-varying function of $t,\lambda_1,L_1,\eta,m,\gamma$ with order $O(\log t)$ on $t$.
        \item  Select the action $x_t$ with the highest upper confidence bound for recommendation.
        \item  Update $\mathbf{Z}_t$,  $\tilde{\mathbf{Z}}_t$ and $\theta_t$ by
        \begin{align*}
            \mathbf{Z}_{t}&=\gamma\mathbf{Z}_{t-1}+\frac{\mathbf{g}(x_{t}; \theta_{t-1}) \mathbf{g}(x_{t} ; \theta_{t-1})^{\top}}{m}\\& +(1-\gamma)\lambda I^{L\times L}.
        \end{align*}
         \begin{align*}
         \tilde{\mathbf{Z}}_{t}&=\gamma\tilde{\mathbf{Z}}_{t-1}+\frac{\mathbf{g}(x_{t} ; \theta_{t-1}) \mathbf{g}(x_{t} ; \theta_{t-1})^{\top}}{m} \\& +(1-\gamma^2)\lambda I^{L\times L}.    
         \end{align*}
         
         Denote the loss function $\mathcal{L}(\theta)$ as
         \begin{align*}
            \mathcal{L}(\theta)=&\sum_{i=1}^{t}\gamma^{-i}(h'(x_{i} ; \theta)-r_{i})^{2} / 2\\&+m \lambda_1\gamma^{-t}\|\theta-\theta^{(0)}\|_{2}^{2} / 2, 
         \end{align*}

 where $0<\gamma<1$ is a discount factor.

 Run $J$ steps of gradient descent on $\mathcal{L}(\theta)$ starting from $\theta_{0},$ and take $\theta_{t}$ as the last iterate:
$$
\theta^{(0)}=\theta_{0}, \theta^{(j+1)}=\theta^{(j)}-\eta \nabla \mathcal{L}(\theta^{(j)}), \theta_{t}=\theta^{(J)}.
$$
\item Update $\gamma_{t}$ by
\begin{align*}
    \gamma_{t}=&O(\underbrace{\sqrt{\lambda_1} S+\nu \sqrt{\log \frac{\operatorname{det}( \tilde{\mathbf{Z}}_{t})}{\delta \operatorname{det} (\gamma^{-t}\lambda_1 I^{L\times L})}}}_{\text {confidence radius }}\\&+\underbrace{err(t,\lambda_1,L_1,\eta,m,\gamma)}_{\text {function approximation error }}).
\end{align*}
Here we leave the calculation of the $\alpha_{t-1}$  and $err$  functions' order on $t,\lambda_1,L_1,\eta,m,\gamma$ as  future work.
    \end{itemize}
\end{itemize}

\subsubsection{Generalization to other online combinatorial optimization problems (O-COP)}\label{othercop}
 
 \textbf{Cascading bandits and position-based bandits.}\label{Cascading}
  In cascading bandits and position-based bandits settings, apart from selecting the item set users prefer most, the relative position of  top-$K$  items also asserts some influence on the aggregate feedback. In these scenarios, we can set the action at each round as a $K\times K$ binary matrix which indicates whether to  place a certain item onto a certain position. For the Wolpertinger sampler and G2ANet sampler, we can first set their outputs as $L$-dimensional real-valued vectors. Then  draw an ordered sample of size $K$ by the Gumbel-Top-$K$ trick \cite{jang2016categorical} and transform the ordered sample into a $K\times K$ binary matrix which can be  
evaluated more easily by the master sampler.

\textbf{GPS Routing, maximum coverage, and other O-COPs.}

  \textbf{(1)} \textbf{$s-t$ shortest path (for GPS routing) 
    \cite{comb}:}
 
         \textbf{(i)}
 Each edge is an arm, and the outcome is the random delay on the
arm from an unknown distribution.

 \textbf{(ii)}  Each  path is a super arm, and the reward is the sum of edge
delays.

 \textbf{(iii)}  Select an $s-t$ path at  each round, and each edge on the path gives
the delay feedback.

 \textbf{(iv)}  Minimize the cumulative delay over all rounds.
 
 \textbf{(2)} \textbf{Maximum coverage \cite{Maximum_coverage_problem}:}
 
     \textbf{(i)} Instance: A number $k$ and a collection of sets $S=\{S_{1}, S_{2}, \ldots, S_{m}\}$.
     
     \textbf{(ii)}  Objective: Find a subset $S' \subseteq S$ of sets, such that $|S^{\prime}| \leq k$ and the number of covered elements $|\bigcup_{S_{i} \in S^{\prime}} S_{i}|$ is maximized.
 
\textbf{(3)}  Etc.

 The above settings can all be modeled as the constrained top-$K$ or cascading CMAB problems.  
In such cases, techniques for top-$K$ or cascading CMAB can be applied and  the semantic loss technique
\cite{xu2018semantic} can be utilized to propagate constraints on an arithmetic circuit in a differentiable manner.

\begin{remark}
Constraint satisfaction is a big challenge when applying deep structures for an end-to-end decision to constrained optimization problems. When encountering numerous and complex constraints, a 
differentiable and unified loss, e.g., semantic loss \cite{xu2018semantic}, is needed to assist  end-to-end training.  By  integrating the  arithmetic circuit that is amenable to backpropagation and the probabilistic sentential decision diagram techniques in constraint programming to capture how close an 
object is to satisfy the constraints,
the  semantic loss function  can specify the separate weight for each dimension of actions and the composite  score of actions' constraint violation condition.  Semantic loss is a very promising tool in constraint satisfaction, but unfortunately due to the lack of   domain knowledge and some implementation errors, we fail to  thoroughly  reproduce this technique in our settings and leave it as  future work.
\end{remark}
 